\documentclass[
submission
]{dmtcs-episciences}

\usepackage[utf8]{inputenc}
\usepackage{subfigure}
\usepackage[numbers]{natbib}
\usepackage{algpseudocode}
\usepackage{algorithmicx,algorithm}

\usepackage{caption}
\usepackage{graphicx}
\usepackage{float} 
\usepackage{subcaption}

\usepackage{booktabs}
\usepackage{multirow}
\usepackage{array, caption, threeparttable}
\usepackage[font=small,labelfont=bf,labelsep=none]{caption}

\author{Yanheng Guo\affiliationmark{1}
  \and Yan Zhang\affiliationmark{1}
  \and Linjie Wu\affiliationmark{1}
  \and Mengxia Li\affiliationmark{1}
  \and Xingjuan Cai\affiliationmark{1,2}\thanks{Corresponding author(xingjuancai@163.com)}
  \and Jinjun Chen\affiliationmark{3}}
\title[Formatting an article for DMTCS]{Edge computing service deployment and task offloading based on multi-task high-dimensional multi-objective optimization}
\affiliation{
  Shanxi Key Laboratory of Big Data Analysis and Parallel Computing, Taiyuan University of Science and Technology, Taiyuan, Shanxi, 030024, China\\
  State Key Laboratory for Novel Software Technology, Nanjing University, P.R. China\\
  Department of Computer Science and Software Engineering, Swinburne University of Technology, Melbourne, Australia}
\begin{document}
\publicationdetails{VOL}{2015}{ISS}{NUM}{SUBM}
\maketitle
\begin{abstract}
  The Mobile Edge Computing (MEC) system located close to the client allows mobile smart devices to offload their computations onto edge servers, enabling them to benefit from low-latency computing services. Both cloud service providers and users seek more comprehensive solutions, necessitating judicious decisions in service deployment and task offloading while balancing multiple objectives. This study investigates service deployment and task offloading challenges in a multi-user environment, framing them as a multi-task high-dimensional multi-objective optimization (MT-HD-MOO) problem within an edge environment. To ensure stable service provisioning, beyond considering latency, energy consumption, and cost as deployment objectives, network reliability is also incorporated. Furthermore, to promote equitable usage of edge servers, load balancing is introduced as a fourth task offloading objective, in addition to latency, energy consumption, and cost. Additionally, this paper designs a MT-HD-MOO algorithm based on a multi-selection strategy to address this model and its solution. By employing diverse selection strategies, an environment selection strategy pool is established to enhance population diversity within the high-dimensional objective space. Ultimately, the algorithm's effectiveness is verified through simulation experiments.
\end{abstract}




\section{Introduction}
The number of Internet users is increasing dramatically along with the rapid development of cloud computing, the Internet of Things (IoT), and mobile terminals. As a result, more businesses are adopting cloud computing technology since it offers safe and manageable access options \cite{1}. The practicality, dependability, and latency of the mobile Internet are under increased pressure to satisfy a variety of service requirements in a variety of circumstances \cite{2,3}. MEC is a revolutionary computational paradigm that intends to bring computing and data storage capabilities close to people and devices at the network edge. To satisfy the increasing needs of mobile apps, the primary goal is to reduce latency and improve compute performance inside mobile networks \cite{4}. Applications may reply more quickly and carry out data processing and storage operations closer to users thanks to MEC's placement of computing resources on edge servers near to consumers. This proximity-driven computing and data processing improves user experiences, lowers network latency, and simultaneously lightens the burden on centralized data centers \cite{5}. Smart cities, IoT, augmented reality (AR), virtual reality (VR), and intelligent transportation systems are just a few of the application scenarios that MEC supports \cite{6}. MEC provides these applications with improved speed and decreased latency through real-time data processing, analysis, and storage on edge servers. According to the distinct characteristics of various applications, MEC's judicious service deployment may optimize the allocation of computing resources, lowering costs and energy usage \cite{7}. Additionally, task offloading allows for the transfer of computationally demanding jobs and data-intensive procedures to edge servers, reducing the strain on mobile devices \cite{8}. As a result, service deployment and task offloading are crucial components of MEC, helping to maximize resource usage, improve system performance, reduce latency, and deliver better user experiences.

A well-planned service deployment may efficiently optimize resource consumption, boost system performance, and increase productivity in edge computing environments. The latency associated with data transmission can be reduced by placing services on edge servers close to consumers, resulting in speedier response times and better user experiences \cite{9}. Maximizing the use of the computing and storage resources available on edge servers is one of the main goals of service deployment. To balance resource use and improve resource utilization efficiency, different service types must be wisely distributed between edge servers \cite{10}. Effective service deployment reduces data transmission latency, which results in quicker reaction times. Numerous initiatives have been made to improve service placement tactics in order to reduce latency while retaining positive user experiences. Additionally, a key component of service deployment research is multi-objective optimization, which aims to balance many goals throughout the service placement process, such as latency, energy consumption, and resource use. An ideal balance between various objectives can be reached by developing suitable trade-off techniques \cite{11}.

In dispersed contexts, task offloading tries to intelligently move tasks from mobile devices to better edge servers or cloud platforms in order to maximize performance and make optimum use of computing resources \cite{12}. MEC is made to give consumers access to services that have lower latency. Allocating user jobs to various servers is a sensible task offloading option that effectively lowers task execution delay. When choosing which tasks to outsource, budget-conscious consumers frequently prioritize task execution costs \cite{13}. Offloading jobs to edge servers with cached services relevant to those activities can reduce data transfer between users and servers, effectively lowering perceived latency for consumers. Different tasks have different service requirements. As a result, service rollout and job offloading must be optimized together \cite{14}. The user experience can be improved by achieving faster response times by caching frequently used services and data on edge servers close to users \cite{15}.

The issues of service deployment and job offloading in edge contexts have been the focus of the aforementioned research projects. Reduced delay, energy savings, and cost savings are only a few advantages of these two components' simultaneous optimization. However, current joint optimization techniques frequently ignore the unique traits of these particular problems. In order to provide stable and reliable services for tasks, service deployment decisions must carefully examine the influence of network circumstances in order to assure network dependability. While simultaneously lowering task execution delay and device energy consumption, task offloading optimization requires obtaining a more equitable distribution of jobs on edge servers. Thus, it becomes very important to ensure load balance while making job offloading decisions.

The distinctive traits of task offloading and service deployment are carefully taken into account in this study. Separate high-dimensional multi-objective optimization models are built for service rollout and task offloading, with a focus on network stability and load balancing, taking latency, cost, and energy usage into account. In the end, both models are concurrently optimized using a multi-task technique, combining them into a single MT-HD-MOO model. In addition, a MT-HD-MOO algorithm based on several selection strategies is suggested for resolving the given problems.

The main contributions of our work comprise:
\begin{itemize} \item A MT-HD-MOO model is constructed to improve the optimization efficiency by exploiting the implied similarity between the two problems of service deployment and task offloading in edge environments. \item 	A MT-HD-MOO algorithm based on multiple selection strategies is proposed, which constructs a pool of selection strategies through multiple selection strategies, and is able to increase the population diversity while retaining the elite individuals. 
\end{itemize}

The subsequent sections of this paper are structured as follows. A review of the work related to service deployment and task offloading in edge environments is presented in Section 2. The proposed models and algorithms are described in detail in Sections 3 and 4 respectively. Section 5 contains a detailed account of the experiments carried out, offering insight into the experimental outcomes. Subsequently, Section 6 encompasses the conclusion drawn from the study and outlines prospects for future research.

\section{related work}
Existing research focuses on dealing with service deployment and task offloading in edge environments. Certain researchers examine the determination of server locations where services are deployed in the MEC system as a challenging task, and the main approaches to solving such problems can be categorized into two types of methods: traditional algorithms and heuristic algorithms. Li et al. \cite{16} devised an optimized strategy for placing edge servers in Ultra-Dense Networks (UDNs), aiming to minimize the service provider's expenses while ensuring timely service completion. However, the service latency was not further optimized to improve the user experience. Li et al. \cite{9} proposed effective service deployment decisions to optimize user experience latency by dynamically adjusting the policy based on the type of task and type of available nodes. But the work only considers the user's experience. Qian et al. \cite{17} pointed out a comprehensive service placement algorithm that guarantees the privacy of customer data and gives high-quality service to the customers. There are many efforts to maximize the user experience and ensure the quality of service, simultaneously, however, they should be optimized from the service provider's point of view concerning goals such as energy consumption and deployment costs. Deng et al. \cite{18} proposed an approach for reducing the cost of application deployment as well as improving the average response time of mobile services under resource constraints and performance requirements. Zhou et al. \cite{19} formulated a delay reduction problem across multiple timeslots within the Lyapunov framework, serving as the foundation for dynamically deploying time-sensitive applications. However, the above two people only consider the time cost as an objective, and in reality, energy consumption, cost, and other objectives need to be taken into account. The majority of the related research focuses on optimizing task execution time, cost, and energy consumption e.g., \cite{20,21,22}, and to deploy the service under the condition that it meets the transaction delay requirements. However, all of the above studies are dedicated to solving the service deployment problem in different scenarios in edge environments, but none of them take the problem of task offloading decisions into consideration.

The issue of supplying computational resources to consumers with a better QoS experience can be solved via task offloading in edge computing \cite{23}. Task offloading decisions depend on conditions such as task requirements, network conditions, and user preferences. Offloading computing tasks can improve the performance and capability of mobile devices while reducing their energy and resource consumption. To ensure offloading availability, Zhao et al. \cite{24} and others viewed the offloading decision of multiple tasks as a multi-objective optimization problem(MOOP) and proposed a centralized algorithm, TS-SMOSA, which allows tasks to be assigned to the best nodes. However, the work assumes that edge nodes can perform all types of tasks, which is not realistic. Wang et al. \cite{25} addressed the task offloading decision within the context of real-life scenarios by formulating it as a single-objective optimization problem and employed a genetic algorithm integrated with elite preservation and stochastic strategies. However, the work combines multiple optimization objectives into a single objective, which is not efficient for optimization. To address the above efficiency issues, Han et al. \cite{26} suggested a collaborative computation offloading algorithm involving multiple nodes, where IoT users break down computational tasks and transfer them to various MEC or cloud nodes. You et al. \cite{27} developed a multi-objective optimization framework that accounts for latency, energy usage, and task execution costs. This framework led to the proposal of a task offloading approach using particle swarm optimization. The objective is to shift tasks from edge devices with limited resources to energy-efficient, low-latency edge servers. Arash et al. \cite{28} et al and Cui et al. \cite{29} used bounded majority purpose optimization which in turn minimizes the consumption of the device and solves the latency problem of the task. However, none of the above work considers the edge servers' maximum cache size and the caching of services.

Taking into account both task and service constraints within the realm of edge computing, addressing the combined optimization challenge of service deployment and task offloading represents a prominent area of research focus. This endeavor aims to achieve the dual objectives of carbon emission reduction and the expansion of sustainable computing practices. Teng et al. \cite{30} proposed a heuristic approach to deal with joint optimization of unloading and deployment problems in an integrated manner. However, the study only improves on the objective of energy utilization and the proposed algorithm is not applicable for optimization of multiple objectives. Song et al. \cite{31} introduced a genetic algorithm-based joint optimization approach to minimize both the individual users' response delay and the average delay collectively. However, it is difficult to deal with three or more targets. Simultaneously tackling the intricacies of service deployment and task offloading introduces a heightened level of complexity to the problem, resulting in a lengthier processing time for the algorithm.

The existence of evolutionary algorithms often focuses on solving a single optimization problem at a time \cite{32}, which will often be a single-objective or multi-objective problem \cite{33}, and the characteristics of the optimization problem are diverse \cite{34}. And Multifactor Evolutionary Algorithm is a paradigm of Evolutionary Multitasking (EMT) algorithms that aims to solve multiple optimization tasks simultaneously using a single population. Gupta et al. \cite{35} combined multi-task evolution to combine MOOP to design algorithms that can solve multiple MOOP simultaneously (MOMFEA). MOMFEA outperforms single-task algorithms because there is some correlation between tasks and potential knowledge transfers can be made. Later \cite{36,37} proposed MFEA-AKT, MOMFEA-SADE, a novel MFEA based on Improved Dynamic Decomposition (MFEA/IDD), and a novel multifactorial evolutionary algorithm for MFEA-GHS, respectively, aiming to accelerate the convergence of the algorithms by exploiting the similarity between the tasks \cite{38,39}, but the lack of selection pressure for the problem in a high-dimensional target space still needs to be addressed.

We discover that in edge contexts, the issues of task offloading and service deployment must be resolved concurrently, and the multitasking optimization algorithm can leverage the similarity between tasks to accelerate algorithm convergence. Therefore, we use a multi-task optimization algorithm to solve the two problems simultaneously and propose multiple selection strategies to construct a pool of selection strategies to solve the problem of insufficient selection pressure in high-dimensional objective space.

\section{Multitasking high-dimensional multi-target modeling in edge environments}

\subsection{Service Deployment Model}
In the context of a multi-access point edge computing configuration comprising $N$ access points and $U$ users, it is plausible to view each access point as an edge cloud, and the set of edge clouds is $N{\rm{ = }}\left\{ {{\rm{0}}{\rm{1}} \cdots ,n} \right\}$ , and the set of users is $U = \left\{ {0,1, \cdots ,u} \right\}$ . To simplify the model, assume that every user possesses a task assigned to them at every given time instance and the task count matches the user count. Several edge clouds can serve numerous users, with users being privy solely to the localized details of the edge computing system. They lack the ability to observe overarching system-wide information.The edge computing service deployment model is as follows:

\begin{equation}
    \sum\limits_{i \in N} {x_u^i = 1} 
\end{equation}
\begin{equation}
    \sum\limits_{u \in U} {{r_u}x_u^i \le {\Phi _i}} 
\end{equation}
\begin{equation}
    x_u^i \in \left\{ {0,1} \right\}
\end{equation}

The service deployment decision variable is $x_u^i$ , $x_u^i = 1$  indicates that user $u$  deploys to the edge cloud $i$ , otherwise, $x_u^i = 0$ ; ${r_u}$  indicates the service requirement of user $u$ ; ${\Phi _i}$ indicates the highest service capacity of the $i$  edge cloud. Eq. (1) signifies that each user is exclusively allocated to a single edge cloud, while Eq. (2) establishes that the aggregate count of services deployed within any given edge cloud must not surpass its allocated resource threshold, and Eq. (3) denotes whether or not the user's   services are deployed on the edge cloud $i$ .

The objective functions for edge computing service deployment include deployment latency, energy consumption, network reliability, and deployment cost.

\subsubsection{Service deployment latency}
Edge computing service deployment latency includes communication latency, task processing latency, and transmission latency.

The access point and the service deployment node of the model in this paper may not be the same node, therefore, the communication transmission delay includes all the transmission time between the access point and the service node. The communication delay ${T_{comm}}$ between the access node and the service node is computed using the following formula:

\begin{equation}
    {T_{comm}} = \sum\limits_{u \in U} {\sum\limits_{i \in N} {x_u^i\varphi _u^i} } 
\end{equation}

where  $\varphi _u^i$ represents the communication latency between access point $u$ and the corresponding edge cloud $i$. The computational delay  ${T_{comp}}$ of the model is calculated as follows: 

\begin{equation}
    {T_{comp}} = \sum\limits_{u \in U} {\frac{{{\gamma _u}}}{{\sum\limits_{i \in N} {{C_i}x_u^i} }}} 
\end{equation}

where the user's  $u$ computing demand  ${\gamma _u}$ is denoted by and the CPU processing power of the computing node  $i$ is denoted by ${C_i}$ .

The following formula is used to compute the delay in transmission ${T_{trans}}$ from the user's device to the access point:

\begin{equation}
    {T_{trans}} = \sum\limits_{u \in U} {\frac{{{\gamma _u}}}{{{v_u}}}} 
\end{equation}

where the transmission rate of the user  $u$ is ${v_u}$ . Therefore, the total delay for service deployment is as follows:

\begin{equation}
    {T_{total}} = {T_{comm}} + {T_{comp}} + {T_{trans}}
\end{equation}
\subsubsection{ Service deployment energy consumption}

Task execution energy consumption encompasses the energy used by users for task transmission into the multi-edge computing network and the energy consumed during task execution by the edge cloud. Energy consumption ${E_t}$ calculation formula is as follows:
\begin{equation}
    {E_t} = \sum\limits_{u \in U} {\sum\limits_{j \in N} {x_j^u{e_j}} } 
\end{equation}

where the energy consumption corresponding to the edge computing node $j$ is ${e_j}$ .

\subsubsection{Service deployment costs}
The cost of deployment consists of two main aspects, the cost of equipment and the cost of transmission. When server providers choose to set up edge servers, they must set up the bare minimum of hardware. Therefore, this part is considered a fixed cost. In addition, the dynamic cost of wire transmission is contingent upon the length of the transmission line. Hence, the cost associated with the placement of an edge server can be represented as:

\begin{equation}
    {\mathop{\rm Cos}\nolimits} {t_{dep}} = \sum\limits_{i = 1}^N {\sum\limits_{j = 1}^S {\left( {{f_\alpha } + {f_\beta }} \right)} } 
\end{equation}

where  ${f_\alpha }$ represents the fixed cost of the edge server ${n_i}$, while ${f_\beta }$ signifies the line transmission cost.

\subsubsection{Service deployment network reliability}

Suppose that each edge server's coverage area resembles a garden with a radius ${R_m}$. To ensure effective service deployment, position the base station within the edge server's coverage area, enabling communication between the base station and the edge server. The following formula can be used to indicate whether the base station  ${s_j}$ is within the coverage area of the edge server ${n_i}$ :

\begin{equation}
    \gamma _j^i = \left\{ \begin{array}{l}
1,d\left( {{n_j},{s_j}} \right) \le {R_m}\\
0,{\rm{      }}otherwise
\end{array} \right.
\end{equation}

In the event that the edge server ${n_i}$ associated with the base station ${s_j}$ becomes non-operational, the base station ${s_j}$ has the capability to opt for the closest functioning server it communicates with. This approach guarantees the reliability of service deployment. The associated reliability objective can be formulated in the following manner:

\begin{equation}
    {\mathop{\rm Re}\nolimits}  = \frac{{\sum\nolimits_{j = 1}^S {\sum\nolimits_{i = 1}^N {\gamma _j^i} } }}{{{S^2}}}
\end{equation}

where $d\left( {{n_i},{s_j}} \right)$  represents the distance separating the edge server at index $i$ and the base station at index $j$, and  $S$ denotes the number of base stations.

\subsection{Task offload model}
The approach segments the task into interlinked subtasks and transfers these subtasks onto the edge nodes for processing. The computing task of the $k$ mobile terminal is denoted as ${V_k} = \left\{ {{v_{k,1}},{v_{k,2}}, \cdots {v_{k,n}}} \right\}$ , where  ${v_{k,n}}$ denotes the $n$  subtask of the $k$  mobile terminal (user). Each user task node is denoted by $\left\{ {{b_{k,n}},{c_{k,n}},{d_{k,n}}} \right\}$ , where, ${b_{k,n}}$  indicates the task node's data size, ${c_{k,n}}$  denotes the $CPU$  resources consumed by the task node for local execution, and  ${d_{k,n}}$ denotes the  $CPU$ resources consumed by the task node for execution at the edge computing node. The decision variable  $s_k^n$ is used to indicate the location where the user  $k$'s   task node is executed, where  $s_k^n = 0$ demonstrates that the task node is run locally, and where  $s_k^n = 1$ indicates the edge server is responsible for executing the tasks assigned to it.

The objective functions of the edge computing task offloading model include computational delay, computational energy consumption, load balancing and offloading cost.

\subsubsection{Task offload calculation delay}
The computational latency model encompasses both the local task execution duration and the latency associated with task offloading to the edge server. The local computation time ${T_{local}}$  is computed as follows:

\begin{equation}
    {T_{local}} = \mathop {\max }\limits_{k \in K} \sum\limits_{n \in N} {\frac{{{c_{k,n}}}}{{{F_{k,local}}}}} 
\end{equation}

The computational power  ${F_{k,local}}$ of the local device $k$  is denoted by. After a task node is loaded to the edge server, its latency  ${T_{edge}}$ is:

\begin{equation}
    ET_k^n = ST_k^n{\rm{ + }}Tser_k^n
\end{equation}
\begin{equation}
    Tedg{e_k} = ET_k^{last} - ST_k^1
\end{equation}
\begin{equation}
    {T_{edge}} = \mathop {\max }\limits_{k \in K} Tedg{e_k}
\end{equation}

where  $ET_k^n$ and  $ST_k^n$ represent the finishing and initiation times of task node $n$ associated with the $k$ mobile device, respectively.  $Tser_k^n$ indicates the time at which the task node on the edge computing server was running;  $ET_k^{last}$ indicates the last task node's termination time of the  $k$ mobile terminal; and  $Tedg{e_k}$ denotes the latency of the  $k$ mobile terminal. Therefore, the edge computing delay  ${T_{offload}}$ can be expressed as:

\begin{equation}
    {T_{offload}} = {T_{local}} + {T_{edge}}
\end{equation}

\subsubsection{Task offload energy consumption}
Calculating energy consumption involves local and edge computing energy,  ${E_{offload}}$ calculated as:

\begin{equation}
    {E_{local}} = {T_{local}} \times {P_{local}}
\end{equation}
\begin{equation}
    {E_{edge}} = {T_{edge}} \times {P_{edge}}
\end{equation}
\begin{equation}
    {E_{offload}} = {E_{_{local}}} + {E_{edge}}
\end{equation}

where  ${E_{local}}$ and  ${E_{edge}}$ denote local computing energy and edge server computing energy, respectively; ${P_{local}}$  and  ${P_{edge}}$ denote local computing power and edge server computing power, respectively.

\subsubsection{Task offload cost}
The whole edge environment is mainly to provide services for users, and user satisfaction is an important indicator for the whole network to focus on improvement, and low offloading cost is an indicator that can respond to user expectations, so an offloading cost objective function is proposed here. When calculating the cost, not only the execution cost of different edge servers in unit time should be considered, but also the cost of computing tasks should be considered.

\begin{equation}
    {\mathop{\rm Cos}\nolimits} {t_{offload}} = \sum\limits_{i = 1}^K {{T_{offload}} \times {P_i} + {Q_i}} 
\end{equation}
where  ${P_i}$ refers to the execution cost per unit time of the $i$  edge server, and  ${Q_i}$ refers to the task's transportation expense to the edge server. A decrease in offloading cost directly corresponds to an increase in user satisfaction.

\subsubsection{Task offload load balancing}
Load balancing stands as a crucial approach for achieving optimal resource utilization across the network. The load balancing optimization objective for edge computing task offloading is calculated as follows:

\begin{equation}
    {L_k} = {\lambda _{k,local,cpu}} + {\lambda _{k,edge,cpu}}
\end{equation}
\begin{equation}
    avg\left( {{L_k}} \right) = \frac{{\sum\limits_{k = 1}^{K + L} {{L_k}} }}{{K + L}}
\end{equation}
\begin{equation}
    {L_{offload}} = \sqrt {\frac{{\sum\limits_{k = 1}^{K + L} {{{\left( {{L_k} - avg\left( {{L_k}} \right)} \right)}^2}} }}{{K + L}}} 
\end{equation}

where  ${L_k}$ denotes the load balancing value of the  $k$ resource; ${\lambda _{k,local,cpu}}$  and ${\lambda _{k,edge,cpu}}$  both denote the $CPU$  utilization rate of the $k$  resource; $avg\left( {{L_k}} \right)$  denotes the average load balancing value; and  ${L_{offload}}$ denotes the load balancing value of the whole network.

\subsection{Multi-tasking high-dimensional multi-objective model for edge environment}
In summary, the MT-HD-MOO model for edge environments can be expressed as follows:
\begin{equation}
    T = \arg \min \left\{ {{T_1},{T_2}} \right\}
\end{equation}
\begin{equation}
    {T_1} = \left\{ \begin{array}{l}
f_1^1 = {T_{total}} = {T_{comm}} + {T_{comp}} + {T_{trans}}\\
f_2^1 = {E_t} = \sum\limits_{u \in U} {\sum\limits_{j \in N} {x_j^u \cdot {e_j}} } \\
f_3^1 = {\mathop{\rm Cos}\nolimits} {t_{dep}} = \sum\limits_{i = 1}^N {\sum\limits_{j = 1}^S {\left( {{f_\alpha } + {f_\beta }} \right)} } \\
f_4^1 = {R_e} = \sum\limits_{i = 1}^N {\sum\limits_{j = 1}^S {\frac{{\gamma _j^i}}{{{S^2}}}} } 
\end{array} \right.
\end{equation}
\begin{equation}
{T_2} = \left\{ \begin{array}{l}
f_1^2 = {T_{offload}} = {T_{local}} + {T_{edge}}\\
f_2^2 = {E_{offload}} = {E_{local}} + {E_{edge}}\\
f_3^2 = {\mathop{\rm Cos}\nolimits} {t_{load}} = \sum\limits_{i = 1}^K {{T_{offload}} \times {P_i} + {Q_i}} \\
f_4^2 = {L_{offload}} = \sqrt {\frac{{\sum\nolimits_{K = 1}^{K + L} {{{\left( {{L_K} - avg\left( {{L_K}} \right)} \right)}^2}} }}{{K + L}}} 
\end{array} \right.
\end{equation}

where $T = \left\{ {{T_1},{T_2}} \right\}$  is a set of two optimization problems  ${T_1}$ (Service Deployment Problem) and  ${T_2}$ (Task Offloading Problem) and in each of them, it contains four objectives.

\section{Multi-task high-dimensional multi-objective algorithm based on multi-selection strategy}
In multi-objective multi-task optimization, the performance of algorithms is often influenced by the increase in the number of objectives involved. An excessive number of non-dominated solutions makes it challenging for the algorithm to effectively distinguish Pareto-dominant solutions, resulting in optimization stagnation. To solve the multi-task high-dimensional multi-objective model in the edge environment, a MT-HD-MOO based on multiple selection strategies (MOMFEA-MS) is designed in this section, which can retain the elite individuals in the external archive set by constructing a pool of selection strategies, which further influences the algorithm's evolutionary process and improves the comprehensive performance of the algorithm.

\begin{algorithm}[t]
\caption{MOMFEA-MS}
\hspace*{0.02in} {\bf Input:}
Population size: $N$; The external population $EP$ ; Randomized mating probability $rmp$ . \\
\hspace*{0.02in} {\bf Output:}
The population $R'$ .
\begin{algorithmic}[1]
\State Randomly initialize  $N$ individuals to obtain an initialized population $R$ ;
\State Skill factor, factor rank, and scalar fitness were calculated separately for each individual in the population;
\While{maximum iteration count not reached}
    \State Perform {\bfseries Algorithm 2} to generate the offspring population $C$;
    \For{ each ${C_a}$ in population $C$ }
        \State  Performing a vertical culture transfer algorithm inherits a skill factor using {\bfseries Algorithm 3};
        \State The individual fitness is calculated according to the individual ${C_a}$  skill factor;
    \EndFor
    \State Find the best individuals in the population and store them in $BestPop$ ;
    \State Combining the offspring population and the current population for $R$ ;
    \State  Building a Pool of Environment Selection Policies using {\bfseries Algorithm 4};
    \State Selection of the next generation of populations $R'$ ;
\EndWhile
\end{algorithmic}
\end{algorithm}
Algorithm 1 outlines the algorithmic structure of MOMFEA-MS. Within Algorithm 1, the population is initialized in a uniform search space (line 1), and each individual is evaluated according to a pre-assigned skill factor (line 2). When the maximum iteration limit is not met (line 3), the parent generation performs the selection and mating algorithm for crossover or mutation operations to generate offspring individuals (line 4), and then performs the vertical culture transfer algorithm for each individual in the population to inherit the skill factors from the parent generation for evaluation, thus realizing the knowledge transfer process (lines 5-8). Merge the offspring population and the contemporary population (line 9), update the external archive set according to the constructed selection strategy pool (line 10), perform environmental selection to obtain a new population (line 11), and so forth until the predefined maximum iteration count is attained.
\subsection{Selection and mating}
Assortative mating is a population search mechanism, whereby individuals in the population tend to mate with others who share a similar cultural background, resulting in offspring. For two randomly selected individuals to undergo a crossover operation and generate offspring, they must share the same skill factor or satisfy a pre-defined assortative mating probability $rmp$ that has been set in advance. Alternatively, a crossover operation can still occur to produce offspring. Achieving an appropriate value of $rmp$ effectively balances the expansion and contraction of the search space. Algorithm 2 presents the pseudocode for the selective mating process:
\begin{algorithm}[t]
\caption{Selective Mating}
\hspace*{0.02in} {\bf Input:}
Paternal individuals ${P_a}$  and ${P_b}$ ; Randomized mating probability $rmp$.\\
\hspace*{0.02in} {\bf Output:}
 The offspring individuals  ${C_a}$ and ${C_b}$ .
\begin{algorithmic}[1]
\State Generate a random number $rand \in \left[ {0,1} \right]$  and compute the skill factors  ${\tau _a}$ and  ${\tau _b}$ for ${P_a}$  and ${P_b}$ ;
\If{$\left( {{\tau _a} = {\tau _b}} \right)$ or $\left( {rand < rmp} \right)$}
    \State   ${P_a}$ and ${P_b}$  perform a crossover operation to generate offspring individuals  ${C_a}$ and ${C_b}$ ;
\Else
    \State  ${P_a}$ and ${P_b}$  each perform a mutation operation to generate offspring individuals  ${C_a}$ and ${C_b}$.
\EndIf
\end{algorithmic}
\end{algorithm}

\subsection{Vertical transmission of culture}
In multifactorial inheritance, vertical cultural transmission is a mechanism that can be carried out in conjunction with biological inheritance so that offspring individuals are directly influenced by their parents. Algorithm 3 depicts its pseudocode. As per the stipulations of Algorithm 2, the generated offspring individuals can be classified into two distinct types: i.e., two cases with two parents and only one parent. Vertical culture transfer mechanism means that when the offspring has two parents, the offspring individual's skill factor is inherited from either parent with an equal probability, while when there is only one parent individual, the offspring can directly acquire the skill factor from its parent individual through inheritance. Vertical cultural transfer mechanism can make the descendants imitate the characteristics of their parents so that the descendants can inherit their parents' skill factors, thus realizing knowledge transfer, i.e., the inheritance of the skill factors is accomplished by the fact that a task can be borrowed from the superior solution of the other task in order to achieve a better result for itself.
\begin{algorithm}[t]
\caption{Vertical Cultural Transmission}
\hspace*{0.02in} {\bf Input:}
Offspring individuals  ${C_a}$ with no assigned skill factor.\\
\hspace*{0.02in} {\bf Output:}
 Offspring individuals ${C_a}$  with assigned skill factor.
\begin{algorithmic}[1]
\If{the offspring individual  ${C_a}$ is generated by crossover}
    \State Produce a random number $rand \in \left[ {0,1} \right]$ ;
    \If{$rand < 0.5$}
        \State The offspring individual ${C_a}$  inherits the skill factor of the parent ${P_a}$ ;
    \Else
         \State The offspring individual ${C_a}$  inherits the skill factor of the parent ${P_b}$ ;
    \EndIf
\EndIf
\If{the offspring individual  ${C_a}$ is generated by mutation}
    \State   The offspring individual  ${C_a}$ inherits the skill factor of the parent ${P_a}$  or ${P_a}$ ;
\EndIf
\end{algorithmic}
\end{algorithm}

\subsection{Selecting policy pools}
The selection strategy of the evolutionary algorithm is an important manifestation of its principle of "survival of the fittest", which is also one of the most important parts. Maintaining the diversity of selection strategies and controlling the selection pressure are the key factors to ensure the performance of evolutionary algorithms. However, the appropriateness of the selection strategy is crucial to the performance of the algorithm. Choosing an unsuitable selection strategy can lead the algorithm to converge towards local optimal solutions, impacting search accuracy and overall robustness. Excessive selection pressure, on the other hand, may lead to lower convergence of the algorithm, thus reducing the search efficiency. Therefore, this paper proposes a pool of environmental selection strategies that can effectively retain elite individuals, which can improve the performance of the solution set in several aspects.

Using three different selection strategies, vector corner-based, tournament-based and grid-based, to select $N$  individuals from the initial population $R$  to generate offspring $R1$ , $R2$ , $R3$ , to construct a total number $3N$  of selection pools, and select  $N$ individuals to be saved in the external archive set $A$  to realize the retention of elite individuals. Expanding the selection pool enhances population diversity and boosts the algorithm's capacity for global exploration, all the while preserving the most exceptional individuals.

\subsubsection{Selection strategy based on vector angles}
The vector angle-based selection strategy is a common high-dimensional multi-objective decision-making method, which is usually used for vector selection problems in multi-dimensional spaces. The ultimate choice made by the strategy is the vector that has the best angle with the reference vector. Specifically, two phases make up the vector angle-based selection strategy: firstly, all the candidate vectors are normalized. Secondly, the angle between every potential vector and the baseline vector is then calculated; and the vector with the smallest angle is selected as the final decision.

The representation of each solution's paradigm in the normalized target space is depicted by Eq. (27); in the normalized space, the vector angles of two solutions are calculated as shown in Eq. (28).
\begin{equation}
    norm\left( {{x_j}} \right) \buildrel \Delta \over = \sqrt {\sum\limits_{i = 1}^m {f_i^{'}} {{\left( {{x_j}} \right)}^2}}  
\end{equation}
\begin{equation}
    angle\left( {{x_j},{y_k}} \right) \buildrel \Delta \over = \arccos \left| {\frac{{{F^{'}}\left( {{x_j}} \right) \cdot {F^{'}}\left( {{y_k}} \right)}}{{norm\left( {{x_j}} \right) \cdot norm\left( {{y_k}} \right)}}} \right|
\end{equation}
where $F'\left( {{x_j}} \right) \cdot F'\left( {{y_k}} \right)$  refers to the inner product of two normalized target vectors.
\begin{equation}
    {F^{'}}\left( {{x_j}} \right) \cdot {F^{'}}\left( {{y_k}} \right) = \sum\limits_{i = 1}^m {f_i^{'}\left( {{x_j}} \right) \cdot f_i^{'}\left( {{y_k}} \right)} 
\end{equation}

\subsubsection{Tournament selection strategy}
The tournament selection strategy is a fitness-based selection method. Its core concept is to select individuals with higher fitness values from the parent population for genetic operations in each generation of the evolutionary process. By doing so, this strategy facilitates the opportunity for superior individuals to produce offspring, thus enhancing the overall fitness values and evolutionary efficiency of the entire population.
\subsubsection{Grid-based selection strategy}
The goal of a grid-based selection strategy is to enhance selection pressure and maintain a broad and even distribution of solutions through a grid-based approach. In particular, this approach employs three grid-based metrics—grid ordering, grid crowding distance, and grid coordinate point distance—to assess individual fitness. This distinction aids in the selection of mating partners and the environmental selection process. Also, to avoid overcrowding, this strategy devises an adaptive adjustment strategy to penalize overcrowded individuals. With this selection strategy, the diversity of the population can be better maintained and ensure that certain solutions are not overly favored while making a selection.

The pseudo-code for the pool of environmental selection strategies is shown in Algorithm 4. ${R_1}$ , ${R_2}$ , ${R_3}$ are the children generated by the three different selection strategies. First, the corresponding children are obtained from different selection operators and combined into a new solution set. Then, the non-dominated individuals in the solution set $S$  are put into the external archive set, and finally updated according to the fitness values calculated by the strategies $SDE$ .

\begin{algorithm}[htbp]
\caption{Environment Selection Policy Pool}
\hspace*{0.02in} {\bf Input:}
Population $R$; External archive set $A$.\\
\hspace*{0.02in} {\bf Output:}
External archive set $A'$.
\begin{algorithmic}[1]
\State ${P_1} = $ Based on vector angle selection $\left( R \right)$;
\State ${P_2} = $ Tournament Selection $\left( R \right)$;
\State ${P_3} = $ Grid-based selection $\left( R \right)$;
\State $S = {P_1} \cup {P_2} \cup {P_3}$
\For{ $i = 1$ to $\left| S \right|$ }
    \For{ $j = 1$ to $\left| A \right|$}
        \State  Compare dominance relationships between individuals $i$  and $j$ ;
        \State  Individuals in $S$ who are not dominated are moved to external archive set $A$.
        \If $\left| A \right| > N$ then
            \If $SDE\left( i \right) > SDE\left( j \right)$
                \State   Retention of individuals $i$ ;
            \Else
                \State   Retention of individuals $j$ ;
            \EndIf
        \EndIf
    \EndFor
\EndFor
\end{algorithmic}
\end{algorithm}

\subsection{Algorithm time complexity analysis}
This section provides a detailed examination of the time complexity associated with the MOMFEA-MS algorithm. The algorithm's time complexity mainly comes from the three strategies in the environment selection strategy pool: the vector angle-based selection strategy, the tournament selection strategy, and the lattice-based selection strategy, which all have a time complexity of $O\left( {{N^2}M} \right)$ . In the selection strategy pool, the population size is $3N$ , and the target space dimension is $M$ . Since these three strategies run independently,the selection strategy pool's overall time complexity is $O\left( {3{N^2}M} \right)$ . The size of the external archive set is often correlated with the population size and is modified through the comparison of individuals within the population against the elite individuals stored in the external archive set, so the maximum time complexity can be represented as $O\left( {{N^2}M} \right)$ . In brief, the most unfavorable time complexity of the MOMFEA-MS algorithm equates to $O\left( {{N^2}M} \right)$ . 

\section{Experiment}
In this work, we simulate a multitasking high-dimensional multi-target model in an edge environment. The devised algorithm is resolved and then juxtaposed with other algorithms for comparison, such as NSGA-III \cite{40}, GrEA \cite{41}, RVEA \cite{42}, and EFR-RR \cite{43} in the simulation environment. Furthermore, an analysis of the algorithms' performance is conducted by evaluating the target values of the model.

\subsection{Experimental setup}
In this study, a three-dimensional region with dimensions of 3km x 3km x 3km is employed. Within this space, there exist numerous users and multiple edge servers. In Table 1, the parameter settings are displayed.

\begin{table}
\caption{Experimental parameter setting}\label{tbl1}
\resizebox{\linewidth}{!}{
\begin{tabular}{@{}cclll@{}}
\toprule
Descriptive                                                 & Value            &  &  &  \\ \midrule
Quantity of edge servers $E$                                & 20               &  &  &  \\
Computational resources of edge servers ${R_{\rm{e}}}$      & {[}1.0,3.0{]}GHz &  &  &  \\
Channel bandwidth $B$                                       & 15MHz            &  &  &  \\
Mobile device transmission power ${P_{\rm{i}}}$             & 0.8W             &  &  &  \\
Mobile device computing resources ${R_m}$                   & {[}0.5,1.2{]}GHz &  &  &  \\
Mobile device to edge server transfer rate $v$              & 170kHz           &  &  &  \\
Number of CPU cycles consumed for local computation $f_c^l$ & 1200 Megacycles  &  &  &  \\
Number of CPU cycles consumed by the edge server $f_c^e$    & 4000 Megacycles  &  &  &  \\
Coverage per base station                                   & {[}100,150{]}m   &  &  &  \\ \bottomrule
\end{tabular}
}
\end{table}

\subsection{model solution}
In this section, MOMFEA-MS will be tested under a multi-task multi-objective scheduling model in an edge environment and compared with the above algorithms. Since different algorithms will solve the model with corresponding deployment and offloading schemes, this results in different objective values. The optimal, worst, and average values of the evaluation metrics obtained by solving the model with the five optimization algorithms are shown in Tables 2 and 3.

\begin{table}[ht]
\caption{Values of different algorithms in Task 1 on a Multitasking Many-Objective Model}
\label{tbl2}
\resizebox{\linewidth}{!}{
\begin{tabular}{cccccc}
\hline
 &
  Algorithm &
  \begin{tabular}[c]{@{}c@{}}Minimize\\ Deployment \\ Latency\end{tabular} &
  \begin{tabular}[c]{@{}c@{}}Minimize\\ Energy    \\ Consumption\end{tabular} &
  \begin{tabular}[c]{@{}c@{}}Maximize\\ Network    \\ Reliability\end{tabular} &
  \begin{tabular}[c]{@{}c@{}}Minimize\\ Deployment    \\ Costs\end{tabular} \\ \hline
\multicolumn{1}{c}{\multirow{5}{*}{Optimum value}} &
  \multicolumn{1}{c}{NSGA-III} &
  \multicolumn{1}{c}{4.26E+04} &
  \multicolumn{1}{c}{3.58E-02} &
  \multicolumn{1}{c}{6.47E-02} &
  \multicolumn{1}{c}{6.27E+04} \\ 
\multicolumn{1}{c}{} &
  \multicolumn{1}{c}{GrEA} &
  \multicolumn{1}{c}{4.69E+04} &
  \multicolumn{1}{c}{3.41E-02} &
  \multicolumn{1}{c}{6.35E-02} &
  \multicolumn{1}{c}{6.21E+04} \\ 
\multicolumn{1}{c}{} &
  \multicolumn{1}{c}{RVEA} &
  \multicolumn{1}{c}{4.33E+04} &
  \multicolumn{1}{c}{3.54E-02} &
  \multicolumn{1}{c}{6.39E-02} &
  \multicolumn{1}{c}{6.34E+04} \\ 
\multicolumn{1}{c}{} &
  \multicolumn{1}{c}{EFR-RR} &
  \multicolumn{1}{c}{5.18E+04} &
  \multicolumn{1}{c}{3.38E-02} &
  \multicolumn{1}{c}{6.44E-02} &
  \multicolumn{1}{c}{6.25E+04} \\ 
\multicolumn{1}{c}{} &
  \multicolumn{1}{c}{MaOMFEA-MS} &
  \multicolumn{1}{c}{4.16E+04} &
  \multicolumn{1}{c}{3.35E-02} &
  \multicolumn{1}{c}{6.52E-02} &
  \multicolumn{1}{c}{6.19E+04} \\ \hline
\multicolumn{1}{c}{\multirow{5}{*}{Worst value}} &
  \multicolumn{1}{c}{NSGA-III} &
  \multicolumn{1}{c}{6.23E+04} &
  \multicolumn{1}{c}{4.42E-02} &
  \multicolumn{1}{c}{6.25E-02} &
  \multicolumn{1}{c}{6.67E+04} \\ 
\multicolumn{1}{c}{} &
  \multicolumn{1}{c}{GrEA} &
  \multicolumn{1}{c}{6.84E+04} &
  \multicolumn{1}{c}{3.89E-02} &
  \multicolumn{1}{c}{6.17E-02} &
  \multicolumn{1}{c}{6.56E+04} \\ 
\multicolumn{1}{c}{} &
  \multicolumn{1}{c}{RVEA} &
  \multicolumn{1}{c}{6.57E+04} &
  \multicolumn{1}{c}{4.59E-02} &
  \multicolumn{1}{c}{6.23E-02} &
  \multicolumn{1}{c}{6.73E+04} \\
\multicolumn{1}{c}{} &
  \multicolumn{1}{c}{EFR-RR} &
  \multicolumn{1}{c}{6.98E+04} &
  \multicolumn{1}{c}{3.66E-02} &
  \multicolumn{1}{c}{6.28E-02} &
  \multicolumn{1}{c}{6.55E+04} \\ 
\multicolumn{1}{c}{} &
  \multicolumn{1}{c}{MaOMFEA-MS} &
  \multicolumn{1}{c}{6.33E+04} &
  \multicolumn{1}{c}{3.63E-02} &
  \multicolumn{1}{c}{6.31E-02} &
  \multicolumn{1}{c}{6.58E+04} \\ \hline
\multicolumn{1}{c}{\multirow{5}{*}{Mean value}} &
  \multicolumn{1}{c}{NSGA-III} &
  \multicolumn{1}{c}{5.25E+04} &
  \multicolumn{1}{c}{3.96E-02} &
  \multicolumn{1}{c}{6.39E-02} &
  \multicolumn{1}{c}{6.43E+04} \\  
\multicolumn{1}{c}{} &
  \multicolumn{1}{c}{GrEA} &
  \multicolumn{1}{c}{5.31E+04} &
  \multicolumn{1}{c}{3.65E-02} &
  \multicolumn{1}{c}{6.25E-02} &
  \multicolumn{1}{c}{6.38E+04} \\
\multicolumn{1}{c}{} &
  \multicolumn{1}{c}{RVEA} &
  \multicolumn{1}{c}{6.12E+04} &
  \multicolumn{1}{c}{4.12E-02} &
  \multicolumn{1}{c}{6.33E-02} &
  \multicolumn{1}{c}{6.46E+04} \\ 
\multicolumn{1}{c}{} &
  \multicolumn{1}{c}{EFR-RR} &
  \multicolumn{1}{c}{6.15E+04} &
  \multicolumn{1}{c}{3.49E-02} &
  \multicolumn{1}{c}{6.36E-02} &
  \multicolumn{1}{c}{6.37E+04} \\ 
\multicolumn{1}{c}{} &
  MaOMFEA-MS &
  5.36E+04 &
  3.44E-02 &
  6.41E-02 &
  6.34E+04 \\ \hline
\end{tabular}
}
\end{table}

\begin{table}[!ht]
\caption{Values of different algorithms in Task 2 on a Multitasking Many-Objective Model}\label{tbl3}
\resizebox{\linewidth}{!}{
\begin{tabular}{cccccc}
\hline
 &
  Algorithm &
  \begin{tabular}[c]{@{}c@{}}Minimize\\ Deployment \\ Latency\end{tabular} &
  \begin{tabular}[c]{@{}c@{}}Minimize\\ Energy    \\ Consumption\end{tabular} &
  \begin{tabular}[c]{@{}c@{}}Maximize\\ Network    \\ Reliability\end{tabular} &
  \begin{tabular}[c]{@{}c@{}}Minimize\\ Deployment    \\ Costs\end{tabular} \\ \hline
\multicolumn{1}{c}{\multirow{5}{*}{Optimum value}} &
  \multicolumn{1}{c}{NSGA-III} &
  \multicolumn{1}{c}{4.26E+04} &
  \multicolumn{1}{c}{3.58E-02} &
  \multicolumn{1}{c}{6.47E-02} &
  \multicolumn{1}{c}{6.27E+04} \\ 
\multicolumn{1}{c}{} &
  \multicolumn{1}{c}{GrEA} &
  \multicolumn{1}{c}{4.69E+04} &
  \multicolumn{1}{c}{3.41E-02} &
  \multicolumn{1}{c}{6.35E-02} &
  \multicolumn{1}{c}{6.21E+04} \\ 
\multicolumn{1}{c}{} &
  \multicolumn{1}{c}{RVEA} &
  \multicolumn{1}{c}{4.33E+04} &
  \multicolumn{1}{c}{3.54E-02} &
  \multicolumn{1}{c}{6.39E-02} &
  \multicolumn{1}{c}{6.34E+04} \\ 
\multicolumn{1}{c}{} &
  \multicolumn{1}{c}{EFR-RR} &
  \multicolumn{1}{c}{5.18E+04} &
  \multicolumn{1}{c}{3.38E-02} &
  \multicolumn{1}{c}{6.44E-02} &
  \multicolumn{1}{c}{6.25E+04} \\  
\multicolumn{1}{c}{} &
  \multicolumn{1}{c}{MaOMFEA-MS} &
  \multicolumn{1}{c}{4.16E+04} &
  \multicolumn{1}{c}{3.35E-02} &
  \multicolumn{1}{c}{6.52E-02} &
  \multicolumn{1}{c}{6.19E+04} \\ \hline
\multicolumn{1}{c}{\multirow{5}{*}{Worst value}} &
  \multicolumn{1}{c}{NSGA-III} &
  \multicolumn{1}{c}{6.23E+04} &
  \multicolumn{1}{c}{4.42E-02} &
  \multicolumn{1}{c}{6.25E-02} &
  \multicolumn{1}{c}{6.67E+04} \\ 
\multicolumn{1}{c}{} &
  \multicolumn{1}{c}{GrEA} &
  \multicolumn{1}{c}{6.84E+04} &
  \multicolumn{1}{c}{3.89E-02} &
  \multicolumn{1}{c}{6.17E-02} &
  \multicolumn{1}{c}{6.56E+04} \\ 
\multicolumn{1}{c}{} &
  \multicolumn{1}{c}{RVEA} &
  \multicolumn{1}{c}{6.57E+04} &
  \multicolumn{1}{c}{4.59E-02} &
  \multicolumn{1}{c}{6.23E-02} &
  \multicolumn{1}{c}{6.73E+04} \\ 
\multicolumn{1}{c}{} &
  \multicolumn{1}{c}{EFR-RR} &
  \multicolumn{1}{c}{6.98E+04} &
  \multicolumn{1}{c}{3.66E-02} &
  \multicolumn{1}{c}{6.28E-02} &
  \multicolumn{1}{c}{6.55E+04} \\ 
\multicolumn{1}{c}{} &
  \multicolumn{1}{c}{MaOMFEA-MS} &
  \multicolumn{1}{c}{6.33E+04} &
  \multicolumn{1}{c}{3.63E-02} &
  \multicolumn{1}{c}{6.31E-02} &
  \multicolumn{1}{c}{6.58E+04} \\ \hline
\multicolumn{1}{c}{\multirow{5}{*}{Mean value}} &
  \multicolumn{1}{c}{NSGA-III} &
  \multicolumn{1}{c}{5.25E+04} &
  \multicolumn{1}{c}{3.96E-02} &
  \multicolumn{1}{c}{6.39E-02} &
  \multicolumn{1}{c}{6.43E+04} \\  
\multicolumn{1}{c}{} &
  \multicolumn{1}{c}{GrEA} &
  \multicolumn{1}{c}{5.31E+04} &
  \multicolumn{1}{c}{3.65E-02} &
  \multicolumn{1}{c}{6.25E-02} &
  \multicolumn{1}{c}{6.38E+04} \\ 
\multicolumn{1}{c}{} &
  \multicolumn{1}{c}{RVEA} &
  \multicolumn{1}{c}{6.12E+04} &
  \multicolumn{1}{c}{4.12E-02} &
  \multicolumn{1}{c}{6.33E-02} &
  \multicolumn{1}{c}{6.46E+04} \\ 
\multicolumn{1}{c}{} &
  \multicolumn{1}{c}{EFR-RR} &
  \multicolumn{1}{c}{6.15E+04} &
  \multicolumn{1}{c}{3.49E-02} &
  \multicolumn{1}{c}{6.36E-02} &
  \multicolumn{1}{c}{6.37E+04} \\ 
\multicolumn{1}{c}{} &
  MaOMFEA-MS &
  5.36E+04 &
  3.44E-02 &
  6.41E-02 &
  6.34E+04 \\ \hline
\end{tabular}
}
\end{table}

From the table, it can be seen that in Task 1, the MOMFEA-MS algorithm has the best optimal values for all four objective values, the best worst values for energy consumption and network reliability, and better performance for the three objective values on the average. Although the RVEA algorithm and the NSGA-III algorithm perform better for the worst and average values of the deployment delay objective, respectively, there is no simultaneous balance between algorithmic diversity and convergence. For Task 2, the MOMFEA-MS algorithm has the best values in most cases. This indicates that the pool of selection strategies covered in this chapter maximizes the retention of some elite individuals, which validates the enhanced effectiveness of the algorithm put forth in this work over the other algorithms compared.

To show the distribution of the solutions, this section uses the form of box plots to visualize the solutions of different algorithms on each objective. Specifically, this is shown in Figures 1 and 2. By looking at the median, upper and lower quartiles, minimum and maximum values of the data, it can be seen that the four algorithms have fewer outliers on the two tasks. For convergence to be good, reflected in the graph is that the values of the median line are optimal on each objective function. Taken together, the MOMFEA-MS algorithm performs better on each objective, with the median line mostly outperforming the other algorithms, showing better convergence and diversity.

\begin{figure}
    \centering
    \begin{minipage}{0.34\linewidth}
        \centering
        \includegraphics[width=.9\linewidth]{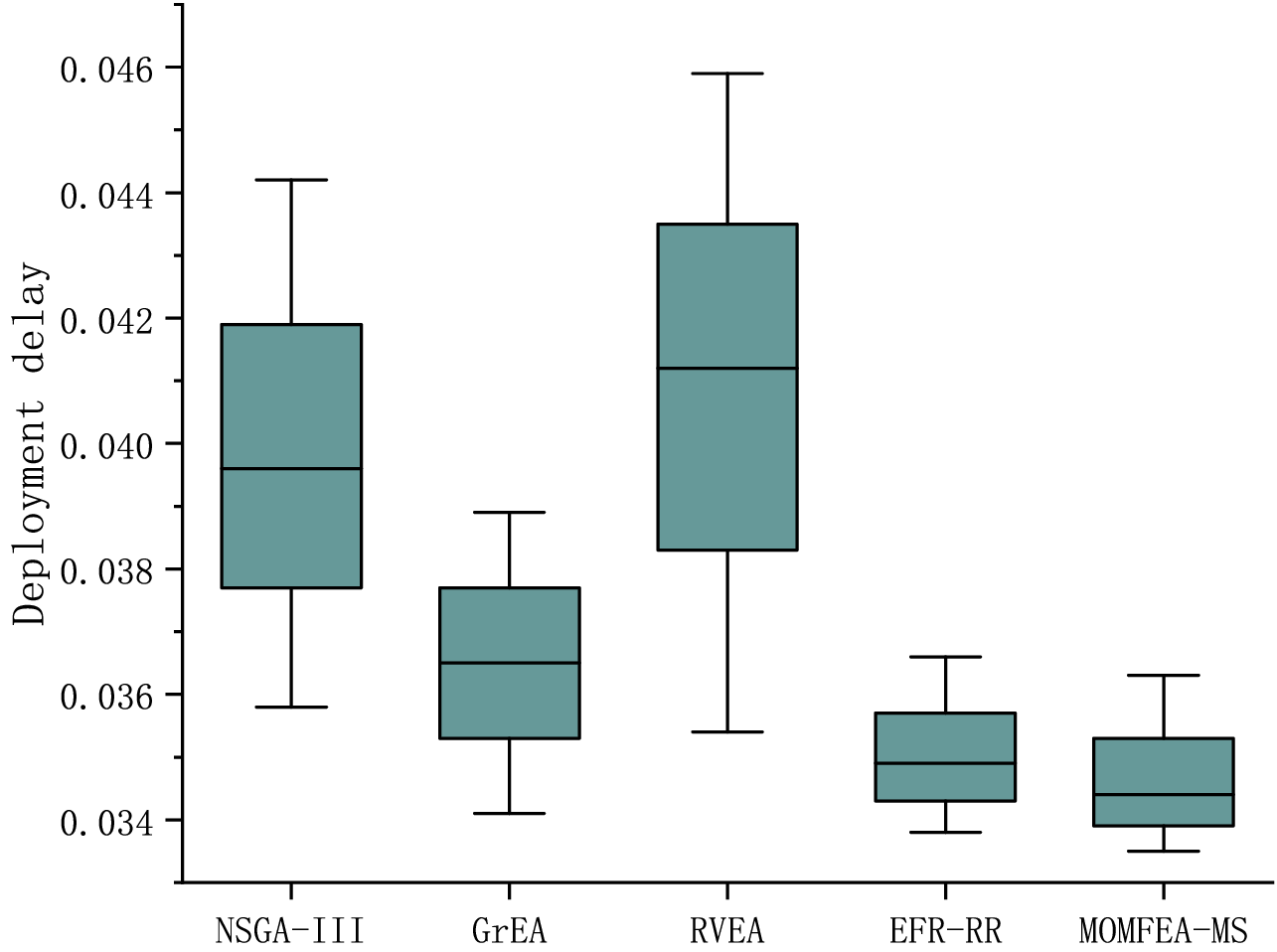}
        \caption*{(a)Task1-Deployment delay}
        \label{Fig11}
    \end{minipage}
    \begin{minipage}{0.34\linewidth}
        \centering
        \includegraphics[width=.9\linewidth]{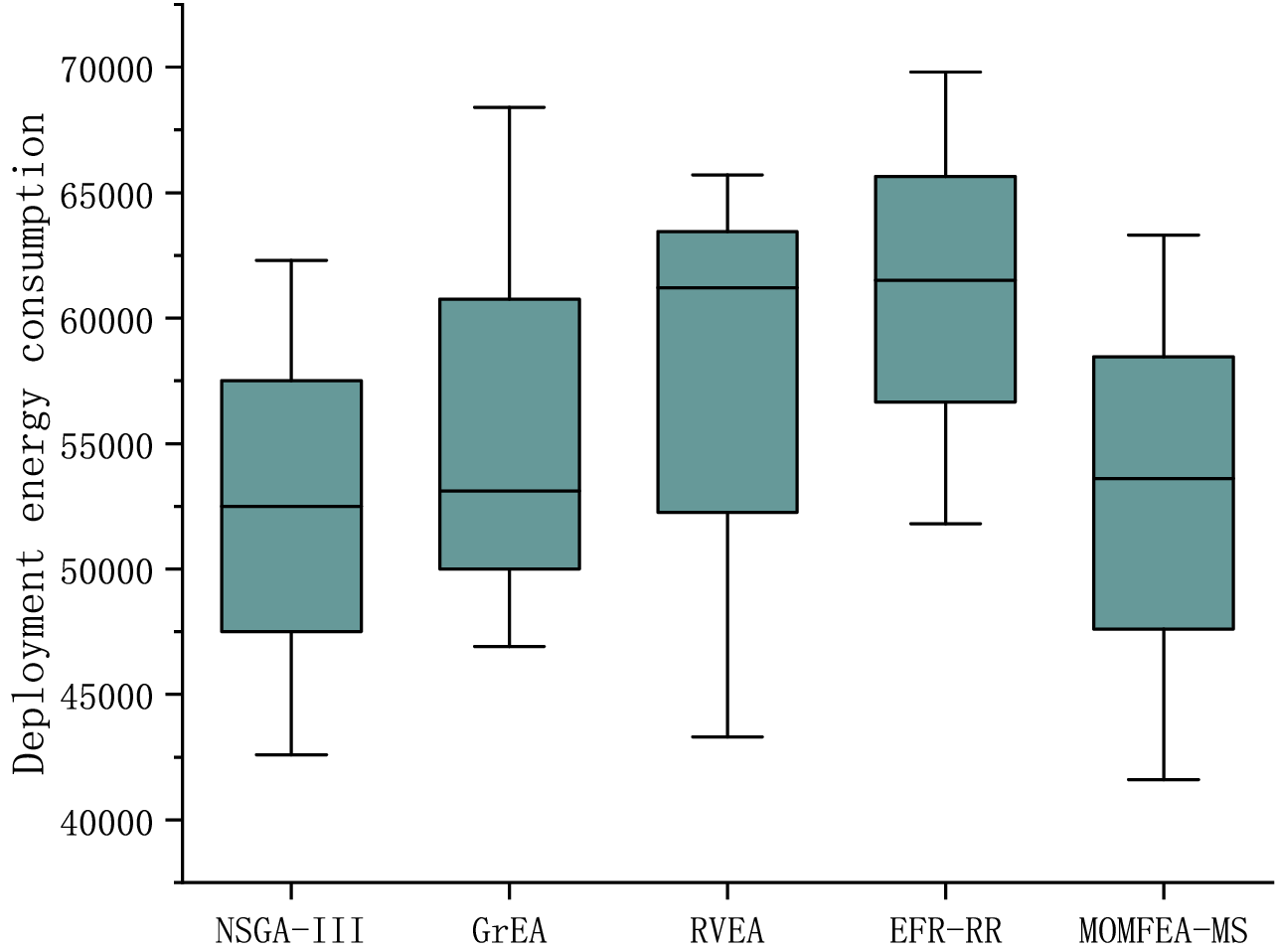}
        \caption*{(b)Task1- Deployment energy}
        \label{Fig12}
    \end{minipage}
    \begin{minipage}{0.34\linewidth}
        \centering
        \includegraphics[width=.9\linewidth]{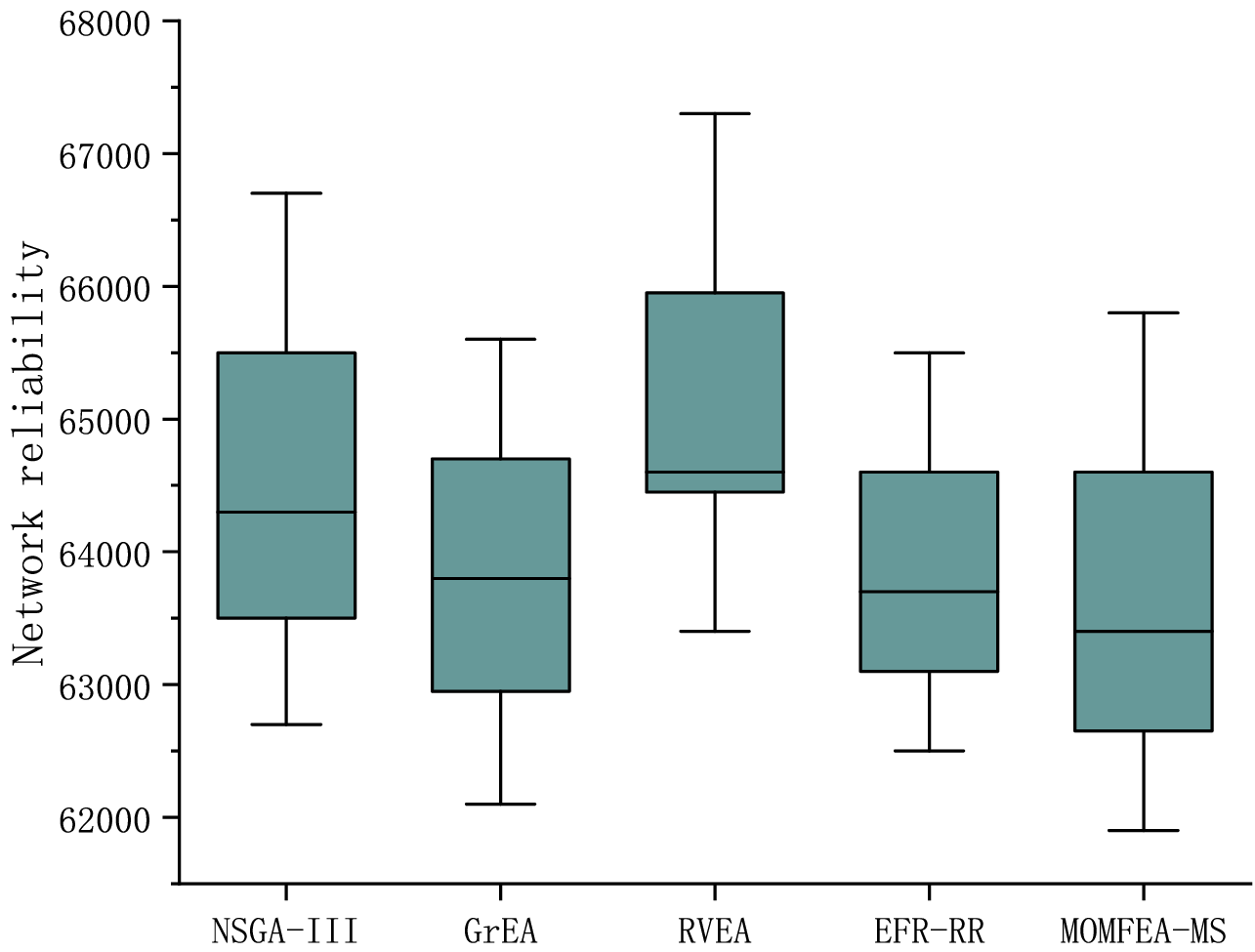}
        \caption*{(c)Task1- Network reliability}
        \label{Fig13}
    \end{minipage}
    \begin{minipage}{0.34\linewidth}
        \centering
        \includegraphics[width=.9\linewidth]{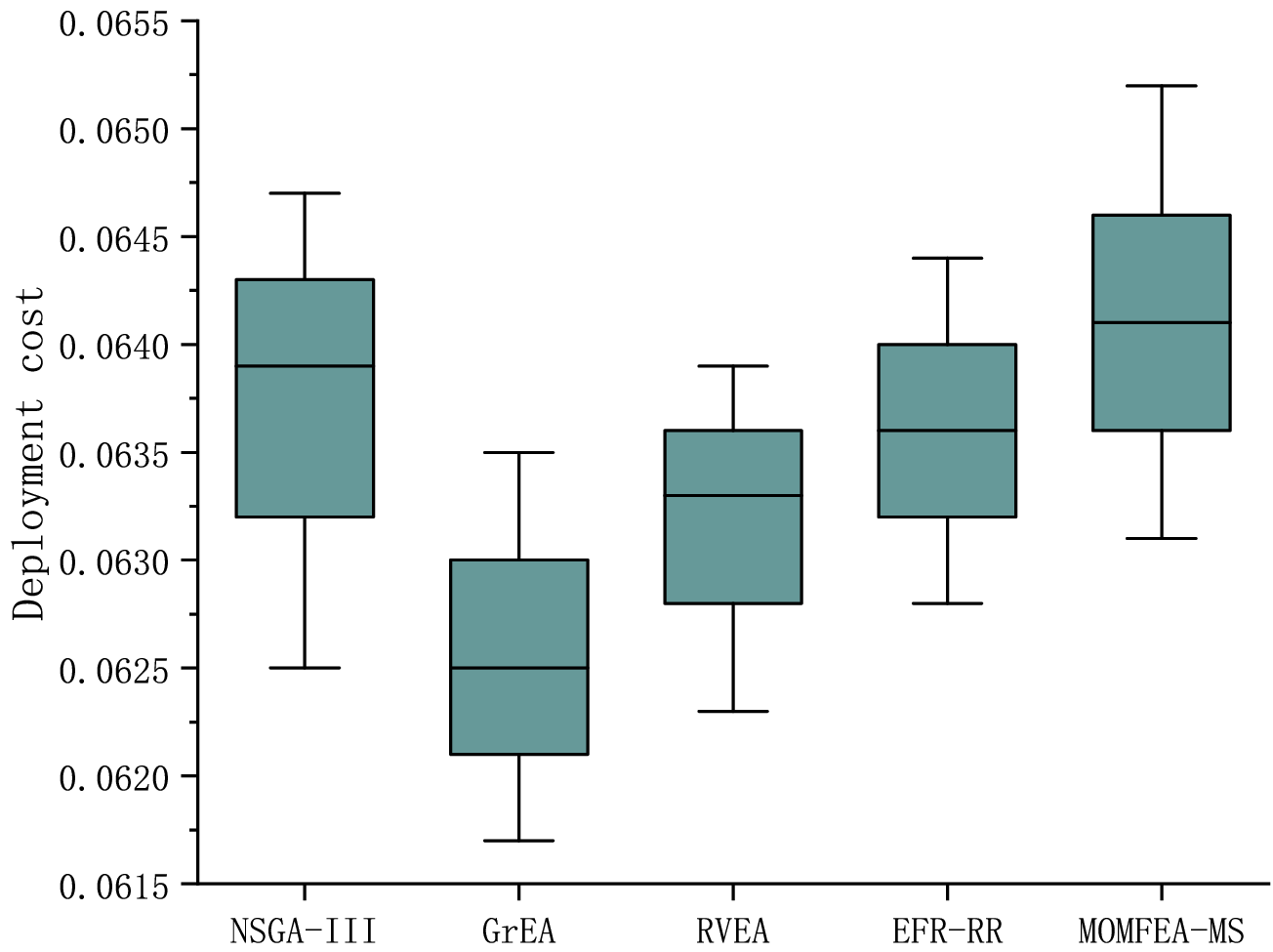}
        \caption*{(d)Task1- Deployment costs}
        \label{Fig14}
    \end{minipage}
    \caption{Performance comparison of different algorithms on the four objectives of Task1}
    \label{Fig1}
\end{figure}
\begin{figure}
    \centering
    \begin{minipage}{0.34\linewidth}
        \centering
        \includegraphics[width=.9\linewidth]{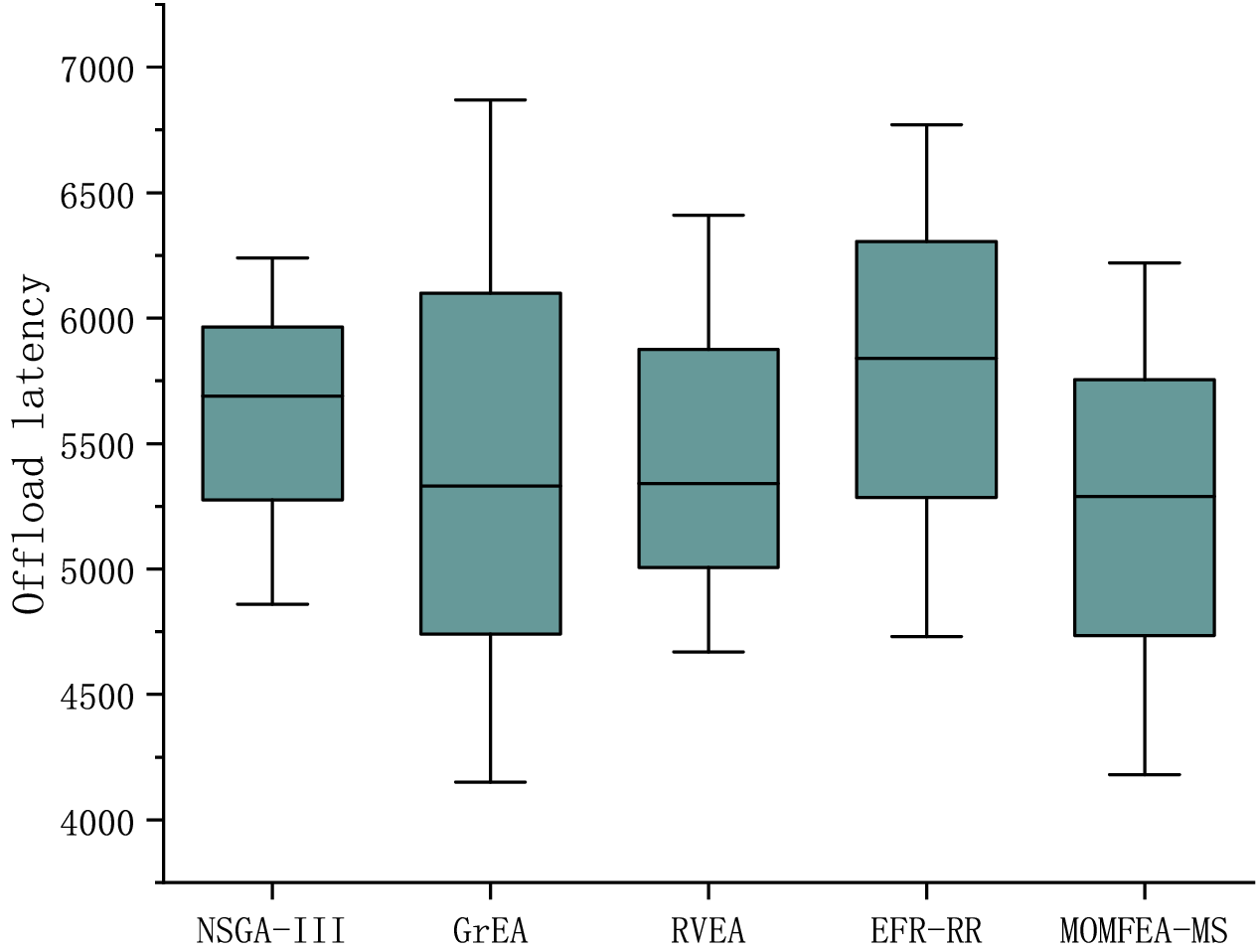}
        \caption*{(a)Task2-Offload latency}
        \label{Fig21}
    \end{minipage}
    \begin{minipage}{0.34\linewidth}
        \centering
        \includegraphics[width=.9\linewidth]{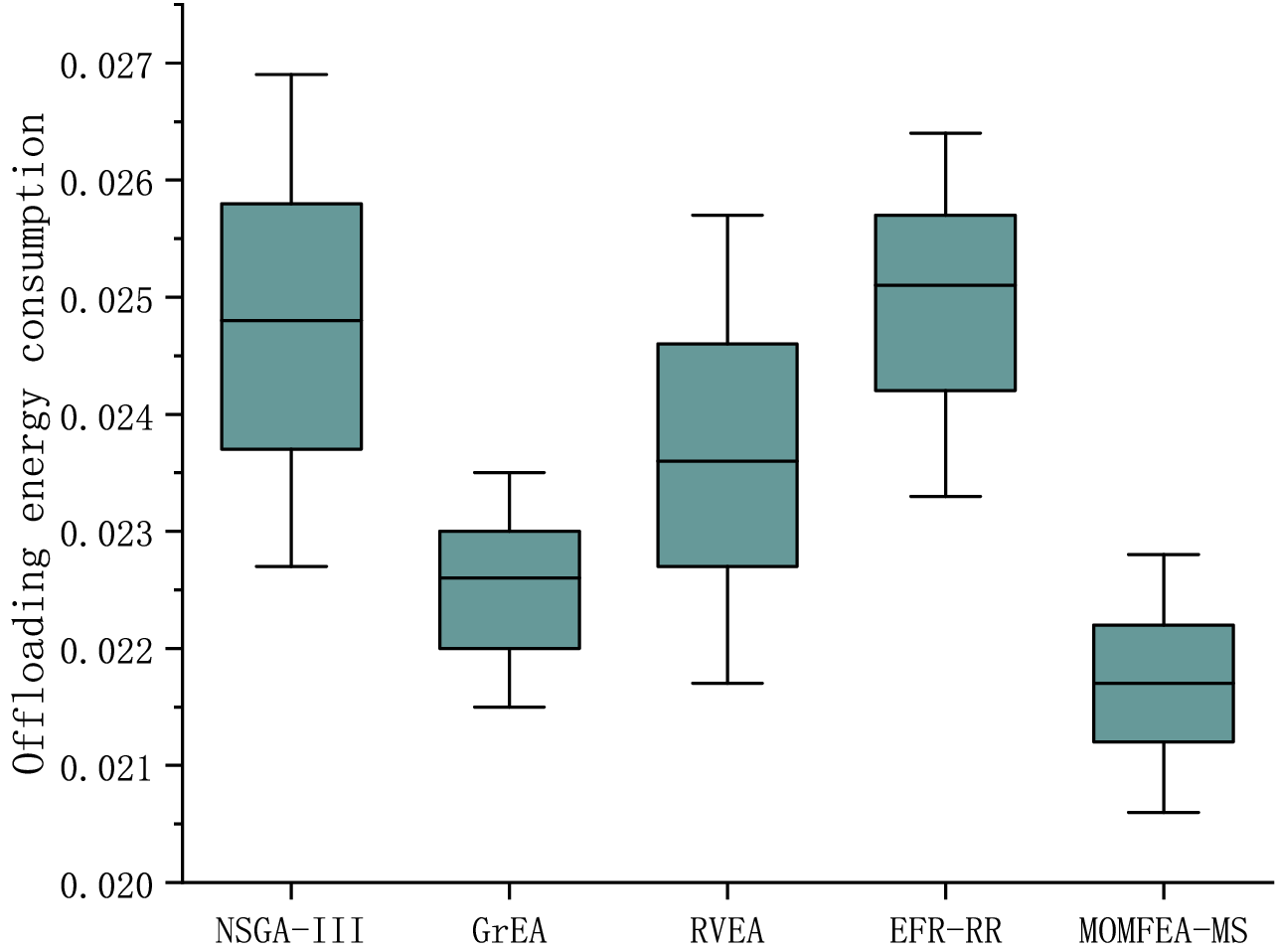}
        \caption*{(b)Task2-Load balancing}
        \label{Fig22}
    \end{minipage}
    \begin{minipage}{0.34\linewidth}
        \centering
        \includegraphics[width=.9\linewidth]{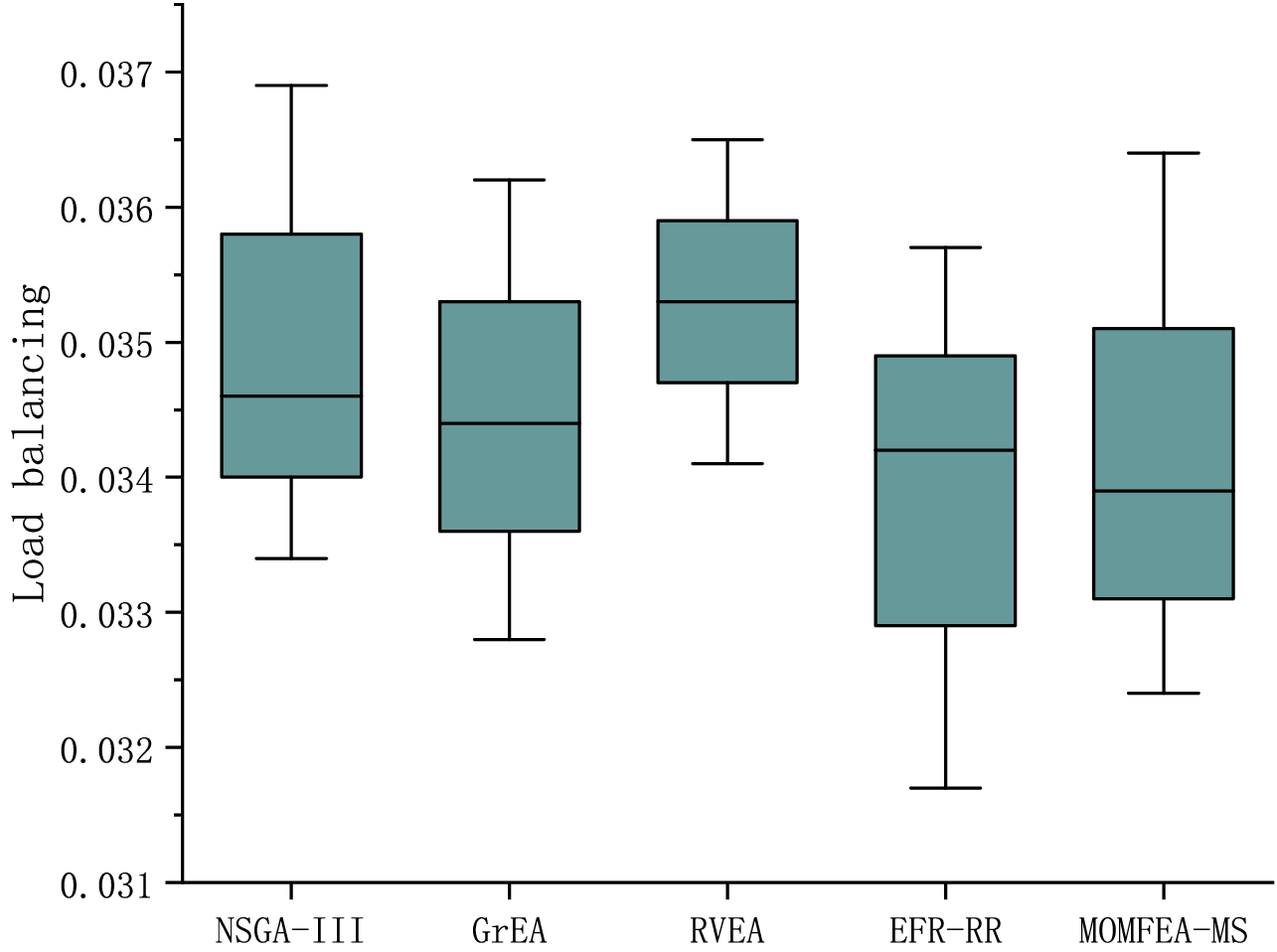}
        \caption*{(c)Task2-Offload energy consumption}
        \label{Fig23}
    \end{minipage}
    \begin{minipage}{0.34\linewidth}
        \centering
        \includegraphics[width=.9\linewidth]{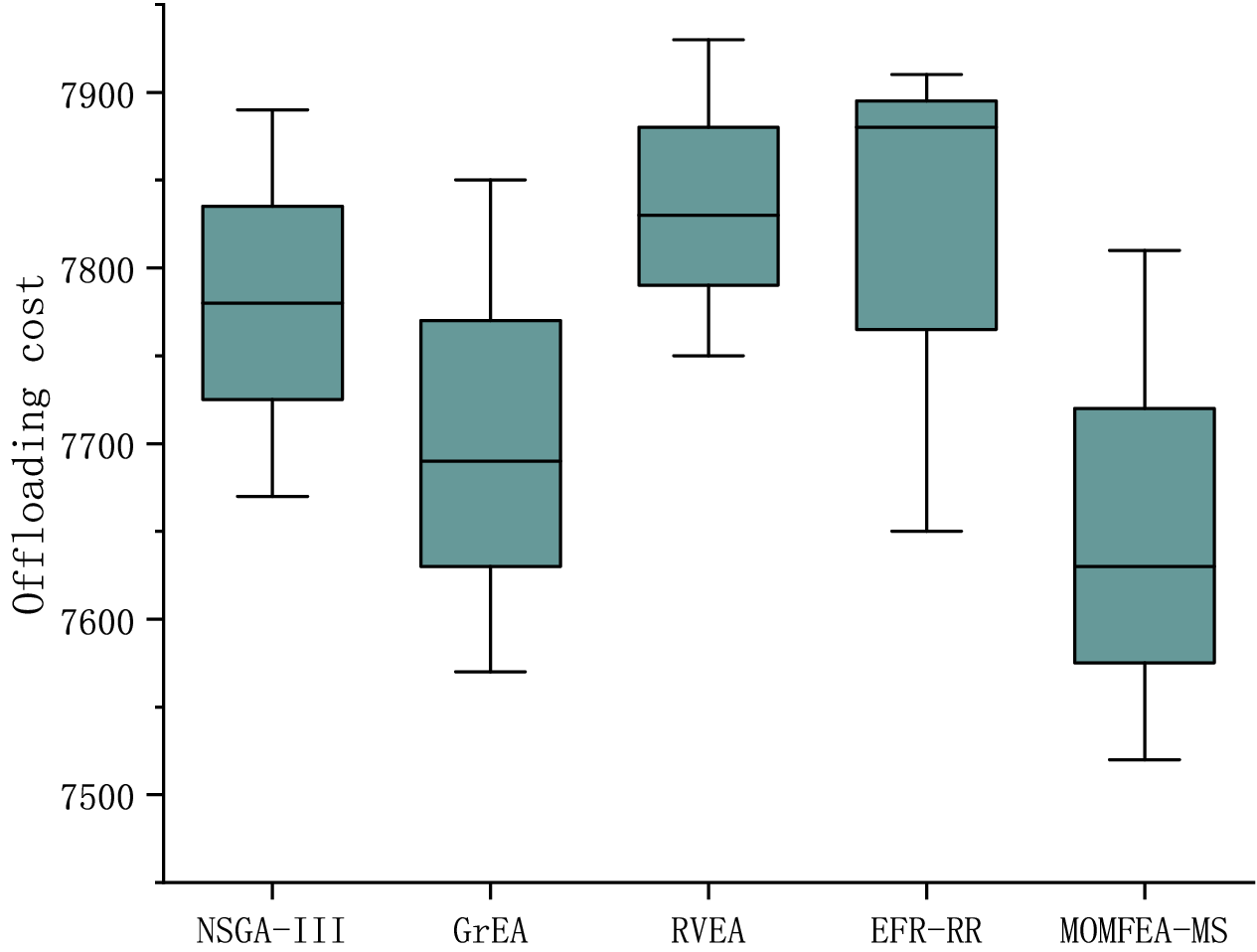}
        \caption*{(d)Task2-Offload cost}
        \label{Fig24}
    \end{minipage}
    \caption{Comparative assessment of various algorithms across the four objectives of Task2}
    \label{Fig2}
\end{figure}
\begin{figure}
    \centering
    \begin{minipage}{0.32\linewidth}
        \centering
        \includegraphics[width=.9\linewidth]{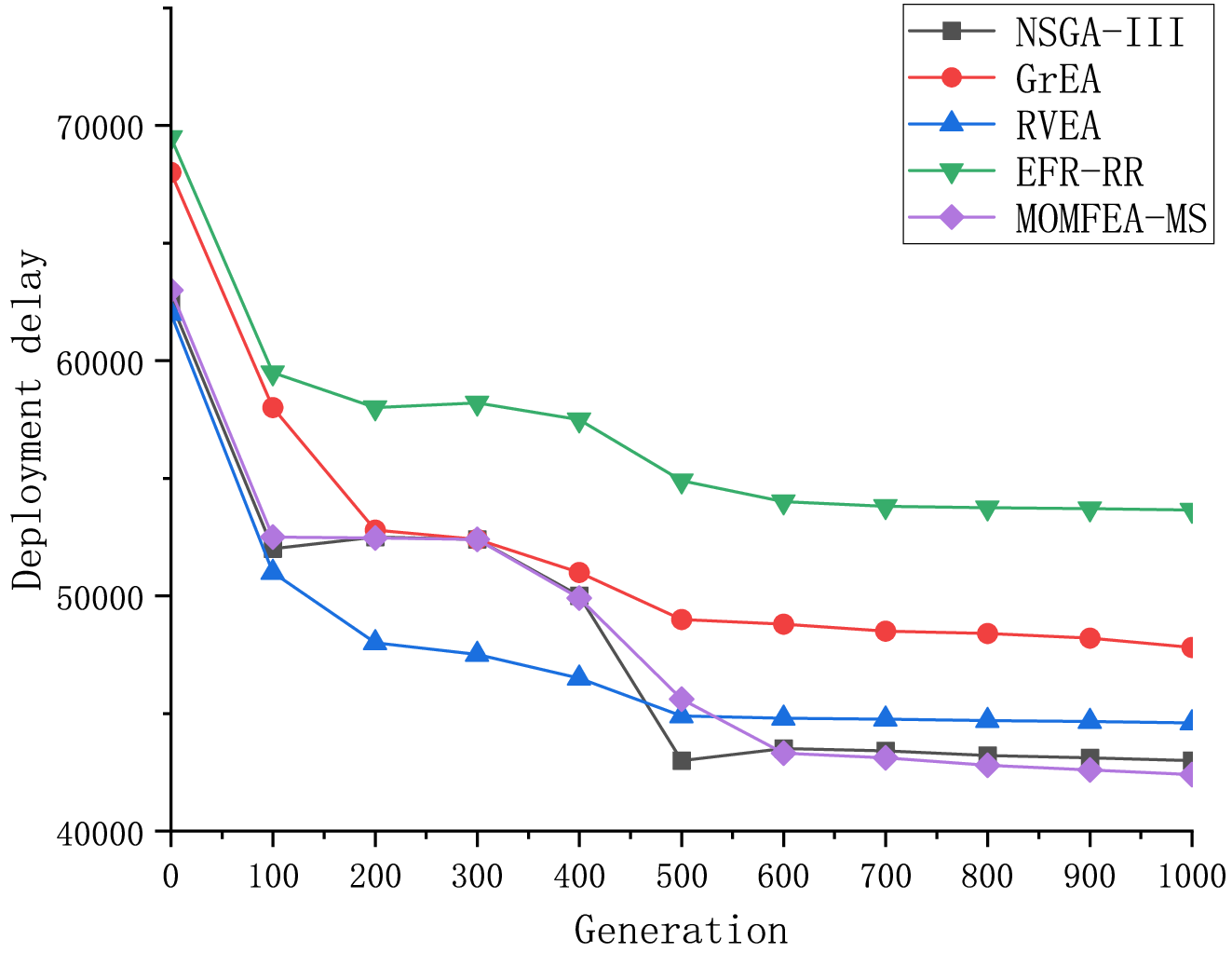}
        \caption*{(a)Task1-Deployment delay}
        \label{Fig31}
    \end{minipage}
    \begin{minipage}{0.32\linewidth}
        \centering
        \includegraphics[width=.9\linewidth]{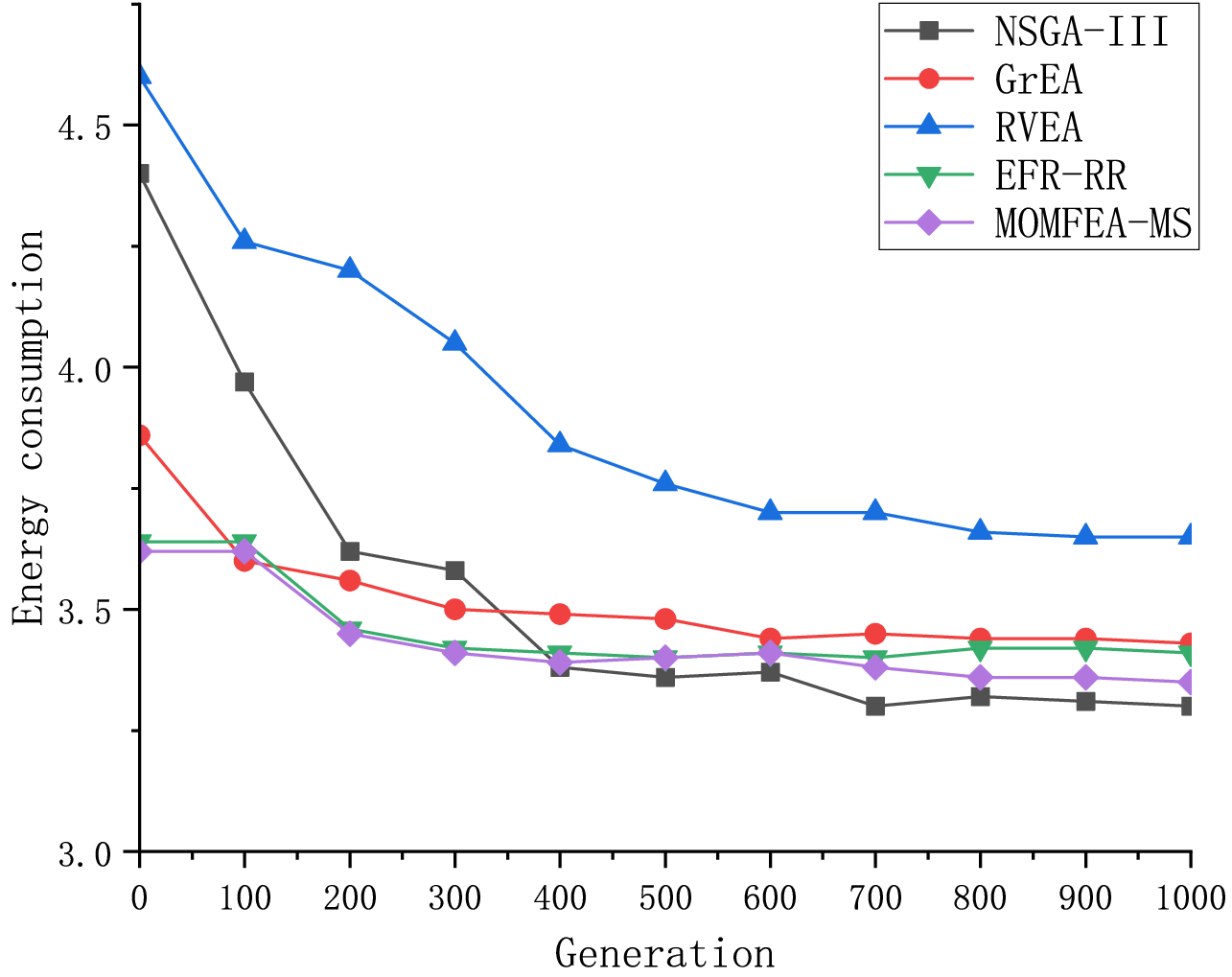}
        \caption*{(b)Task1- Deployment energy}
        \label{Fig32}
    \end{minipage}
    \begin{minipage}{0.32\linewidth}
        \centering
        \includegraphics[width=.9\linewidth]{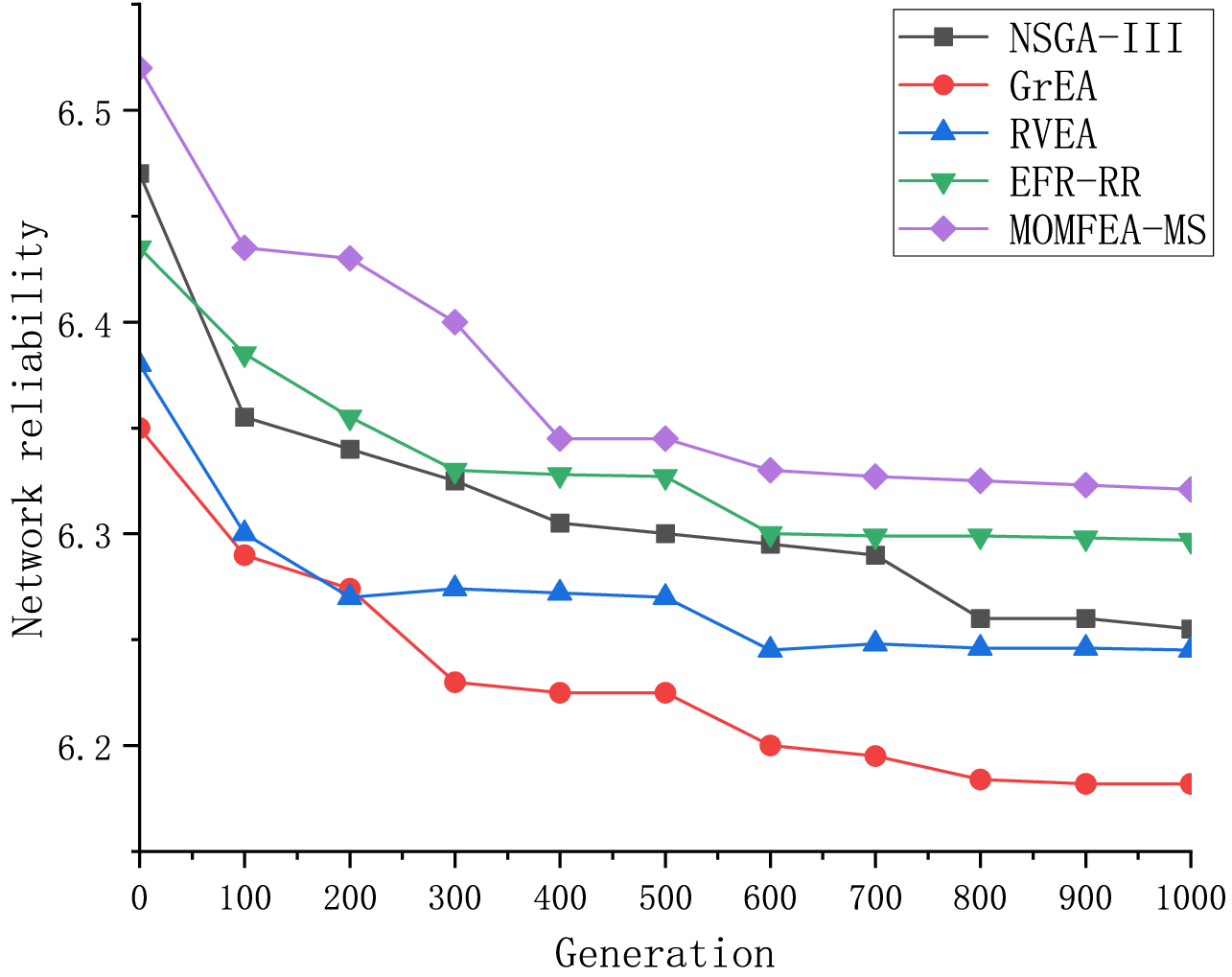}
        \caption*{(c)Task1- Network reliability}
        \label{Fig33}
    \end{minipage}
    \begin{minipage}{0.32\linewidth}
        \centering
        \includegraphics[width=.9\linewidth]{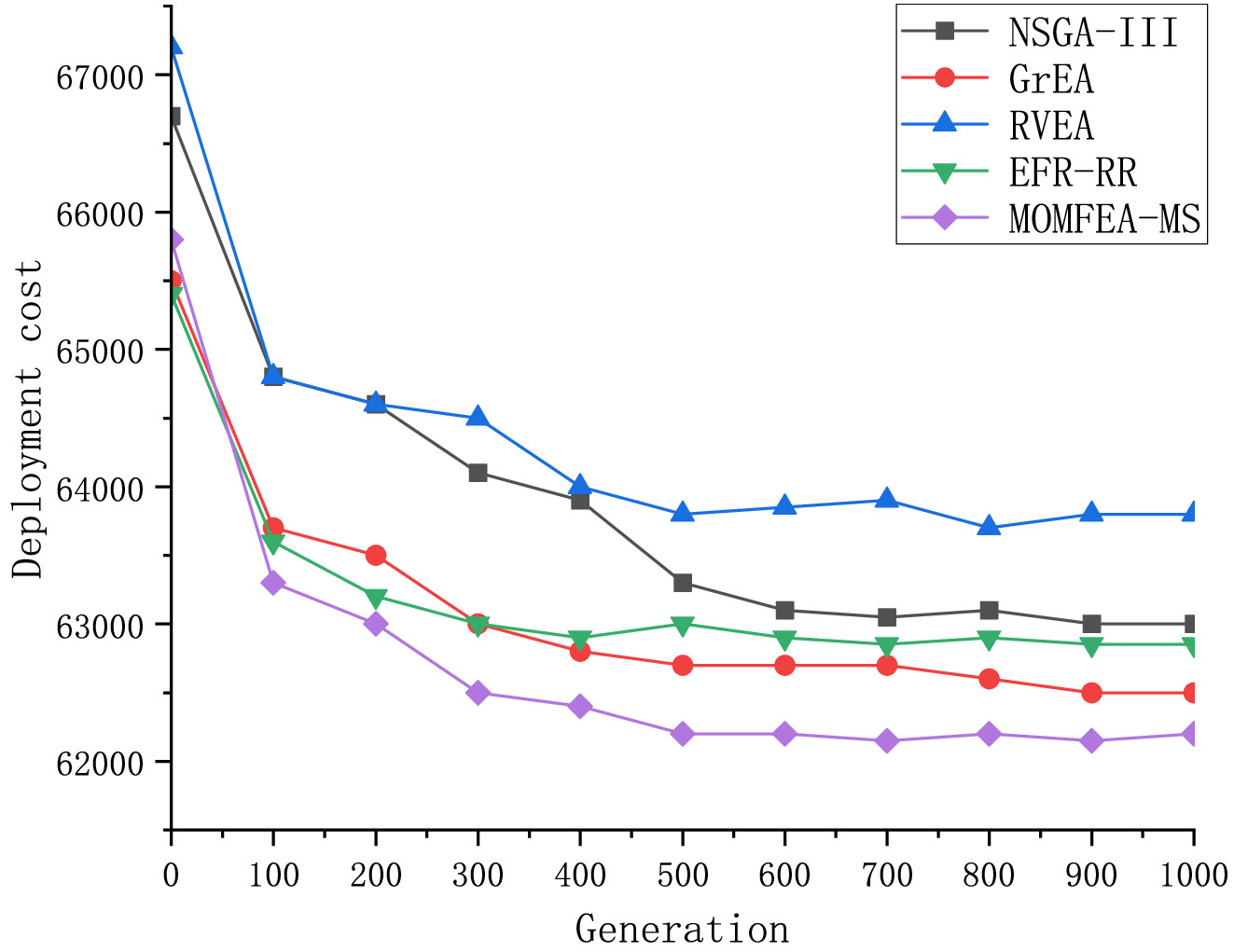}
        \caption*{(d)Task1- Deployment costs}
        \label{Fig34}
    \end{minipage}
    \caption{Effect of different number of iterations on algorithm performance in Task 1}
    \label{Fig3}
\end{figure}
\begin{figure}
    \centering
    \begin{minipage}{0.31\linewidth}
        \centering
        \includegraphics[width=.9\linewidth]{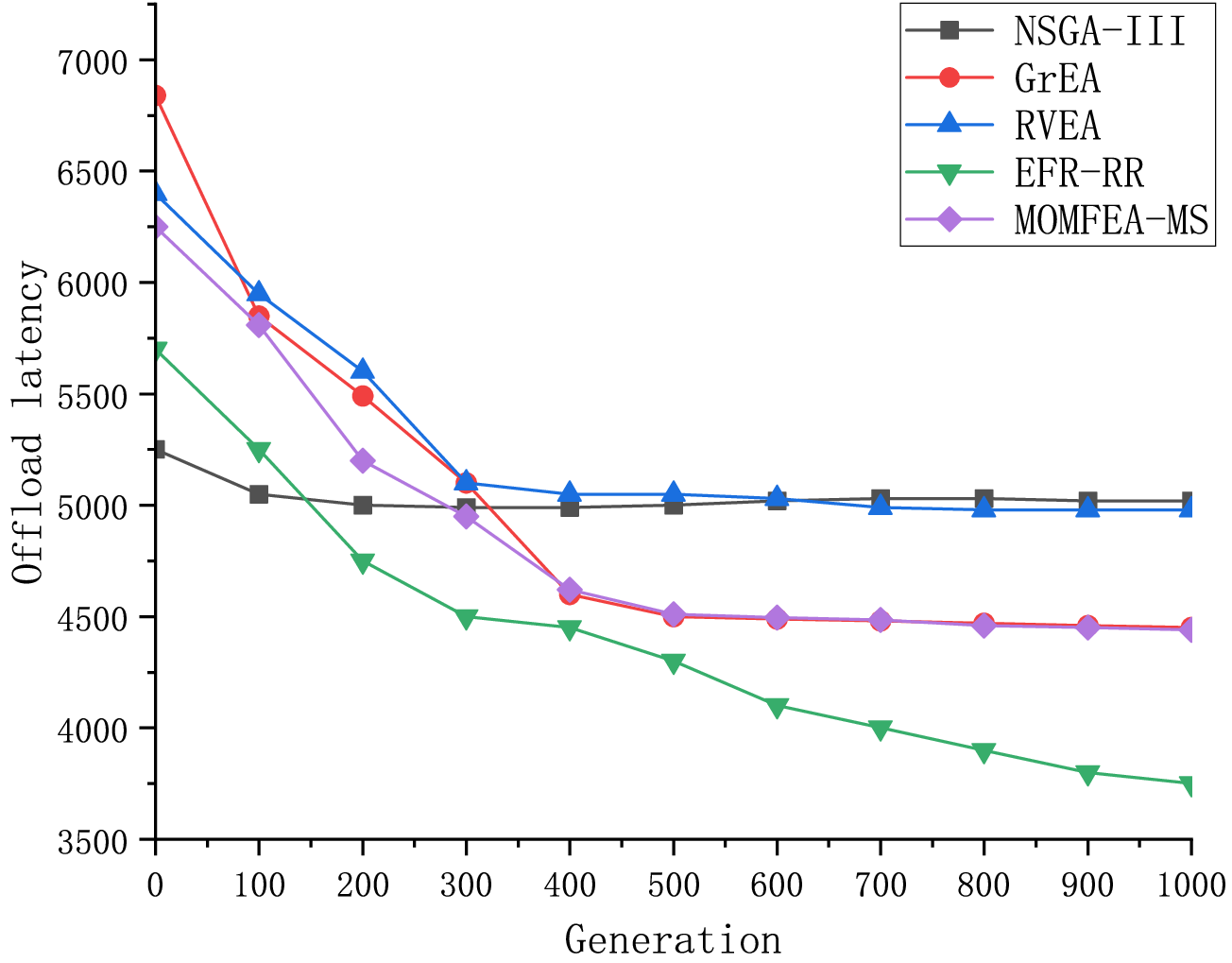}
        \caption*{(a)Task2-Offload latency}
        \label{Fig41}
    \end{minipage}
    \begin{minipage}{0.31\linewidth}
        \centering
        \includegraphics[width=.9\linewidth]{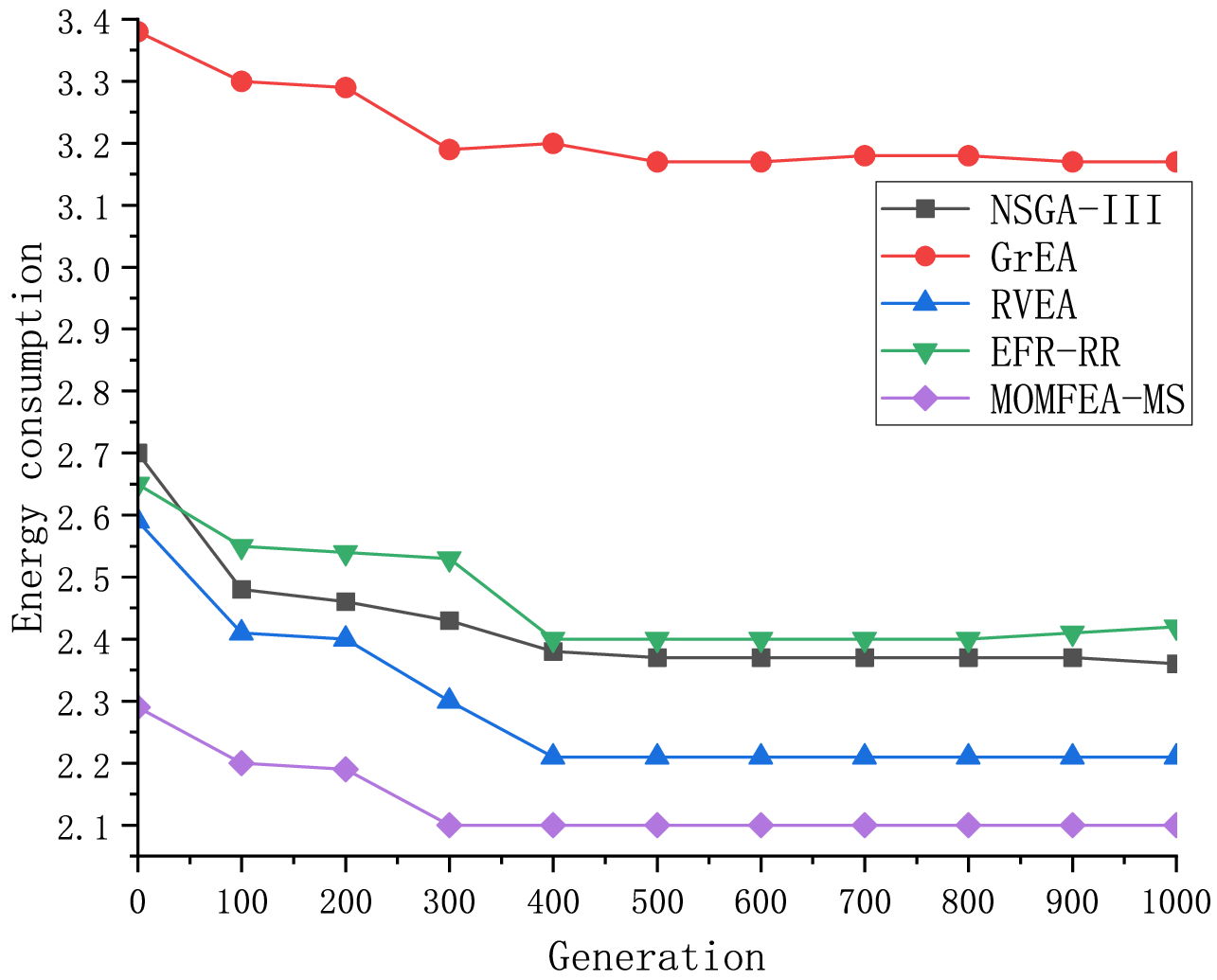}
        \caption*{(b)Task2-Load balancing}
        \label{Fig42}
    \end{minipage}
    \begin{minipage}{0.31\linewidth}
        \centering
        \includegraphics[width=.9\linewidth]{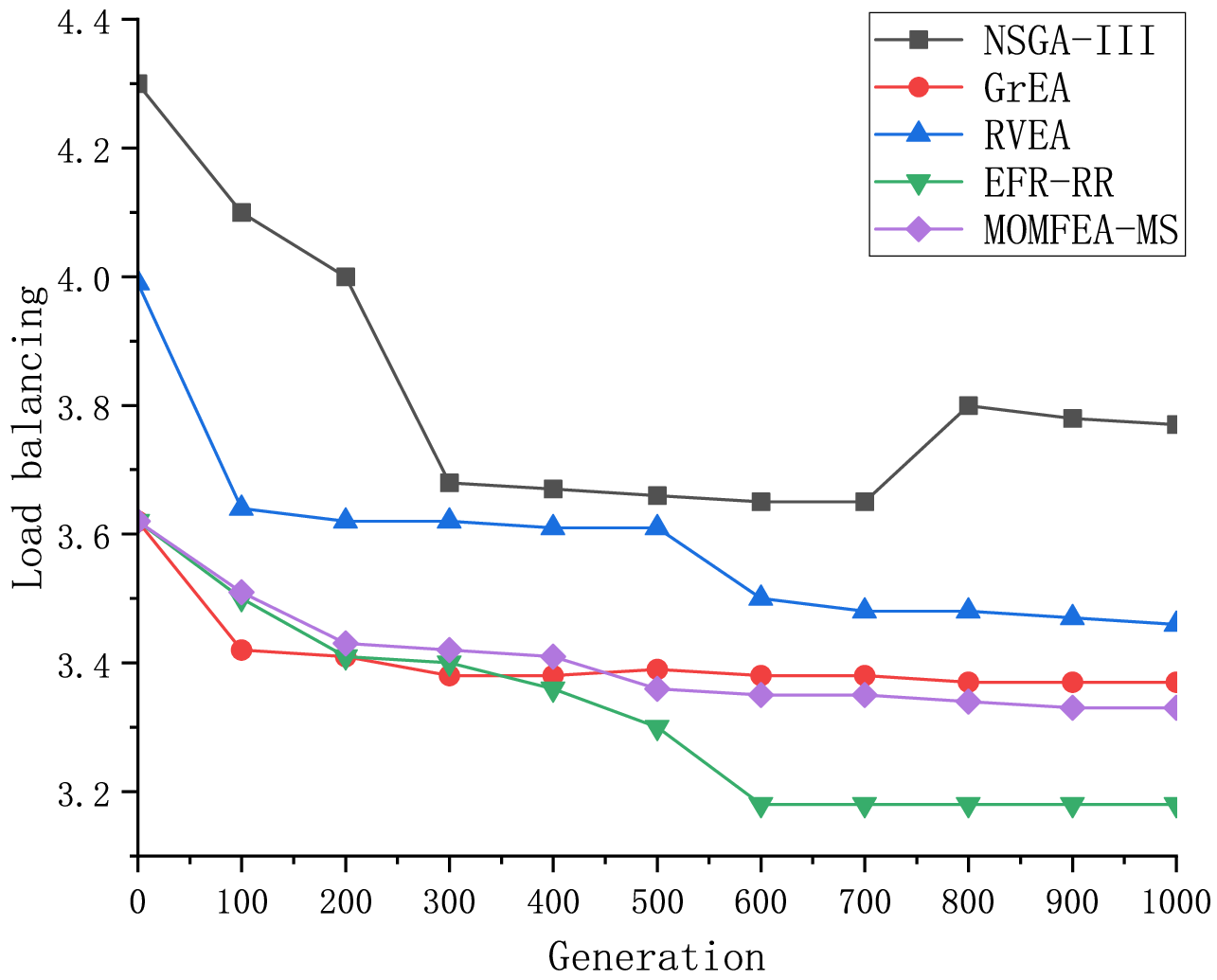}
        \caption*{(c)Task2-Offload energy consumption}
        \label{Fig43}
    \end{minipage}
    \begin{minipage}{0.31\linewidth}
        \centering
        \includegraphics[width=.9\linewidth]{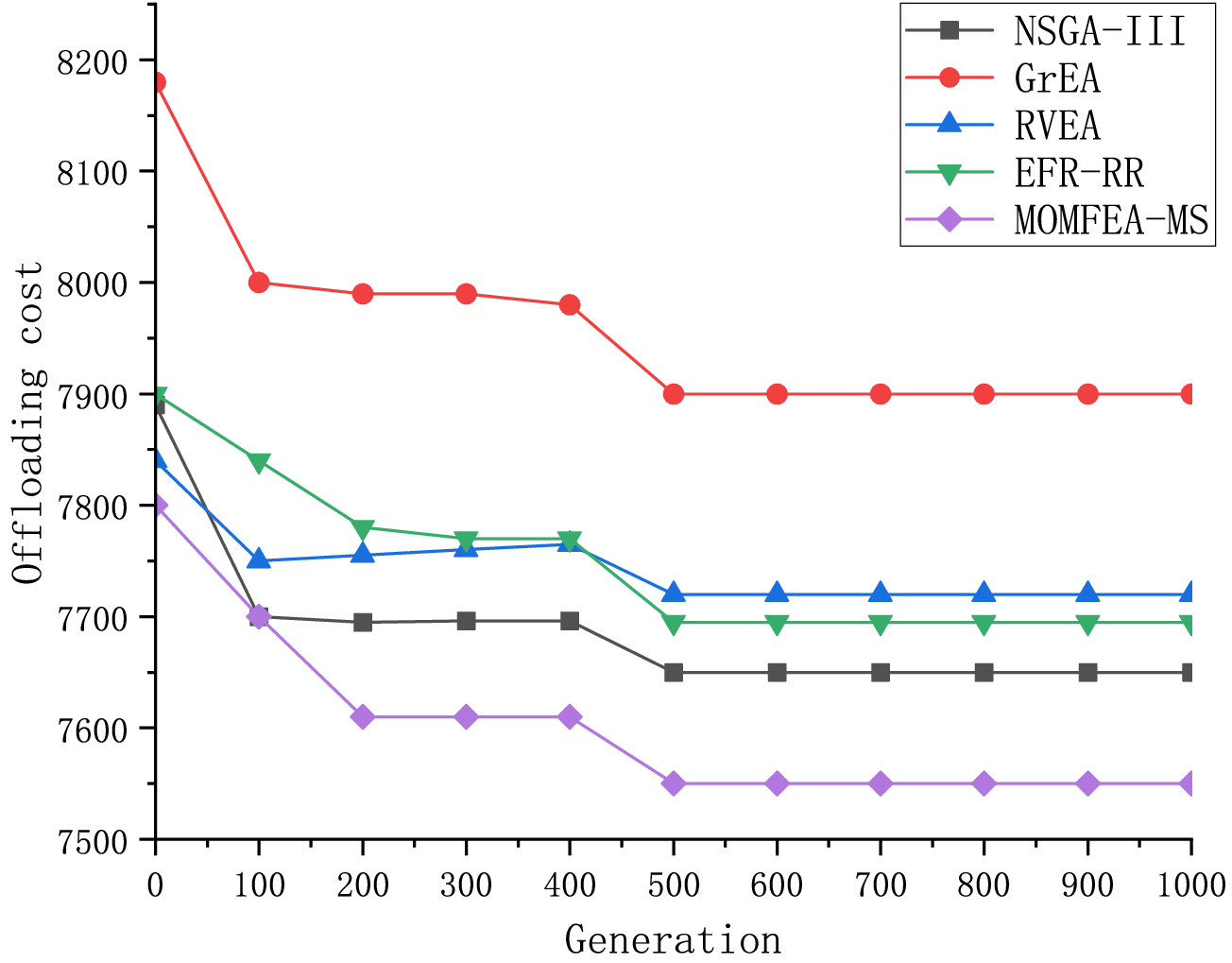}
        \caption*{(d)Task2-Offload cost}
        \label{Fig44}
    \end{minipage}
    \caption{Effect of different number of iterations on algorithm performance in Task 2}
    \label{Fig4}
\end{figure}
Meanwhile, in Figs. 3 and 4, task 1 and task 2 are compared on different objectives of different algorithms, illustrating the influence of the iteration count on the algorithm's performance. The algorithm's swift convergence is noticeable, gradually approaching the optimal value within 100 iterations; whereas, with the increment in the number of iterations, the algorithm's convergence speed decreases significantly, especially between 300 and 600 iterations. Although the algorithm reaches the optimal value of around 600 iterations, there are still slight fluctuations. Overall, the results of the MOMFEA-MS algorithm fluctuated within the range of the optimal value, but the results of the objective value gradually stabilized as the number of iterations increased.

\section{Conclusion}
The peculiarities of task offloading and service deployment in edge contexts are acknowledged in this study. Network reliability and load balancing are emphasized as the main optimization goals in two high-dimensional multi-objective optimization models. In order to concurrently optimize both models, a multi-task technique is used, taking use of the inherent similarities between the two issues. An environment selection strategy pool is built utilizing several selection methods to help the algorithm reach a variety of solutions, taking into account the possible efficiency decrease of optimization algorithms in high-dimensional objective spaces. We will go deeper into the multi-task algorithms' application possibilities in the future work.

\section*{Acknowledgements}

This project was funded by the National Natural Science Foundation of China under Grant No. 61806138the National Key Research and Development Program of China under Grant No. YDZJSX2021A038 the Open Fund of State Key Laboratory for Novel Software Technology (NanjingUniversity) under Grant No. KFKT2022B18. 

\nocite{*}
\bibliographystyle{abbrvnat}
\bibliography{sample-dmtcs}
\label{sec:biblio}

\end{document}